\documentclass[acmsmall]{acmart}
\graphicspath{{figs/}} 
\usepackage{tabularx}
\usepackage{multirow}
\usepackage{booktabs}
\usepackage{subcaption} 
\usepackage{color}
\usepackage{natbib}

\setcitestyle{square, comma, numbers,sort&compress, super}
\AtBeginDocument{%
  \providecommand\BibTeX{{%
    \normalfont B\kern-0.5em{\scshape i\kern-0.25em b}\kern-0.8em\TeX}}}

\setcopyright{acmcopyright}
\copyrightyear{2024}
\acmYear{2024}
\acmDOI{XXXXXXX.XXXXXXX}

\acmJournal{CSUR}
\acmVolume{37}
\acmNumber{4}
\acmArticle{1}
\acmMonth{1}

\begin{document}

\title{Topology-aware Federated Learning in Edge Computing: A Comprehensive Survey}

\author{Jiajun Wu}
\authornotemark[1]
\email{jiajun.wu1@ucalgary.ca}
\orcid{0000-0003-1000-7356}
\affiliation{%
  \institution{Department of Electrical and Software Engineering, University of Calgary}
  \streetaddress{2500 University Drive NW}
  \city{Calgary}
  \state{Alberta}
  \country{Canada}
  \postcode{T2N 1N4}
}

\author{Fan Dong}
\affiliation{%
  \institution{Department of Electrical and Software Engineering, University of Calgary}
  \streetaddress{2500 University Drive NW}
  \city{Calgary}
  \state{Alberta}
  \country{Canada}}
\email{fan.dong@ucalgary.ca}
\orcid{0009-0005-0262-791X}

\author{Henry Leung}
\affiliation{%
  \institution{Department of Electrical and Software Engineering, University of Calgary}
  \streetaddress{2500 University Drive NW}
  \city{Calgary}
  \state{Alberta}
  \country{Canada}}
\email{leungh@ucalgary.ca}
\orcid{0000-0002-5984-107X}

\author{Zhuangdi Zhu}
\affiliation{%
  \institution{Department of Computer Science and Engineering, Michigan State University}
  \streetaddress{428 S Shaw Ln Rm 3115, East Lansing, MI 48824}
  \city{East Lansing}
  \country{USA}}
\email{zhuzhuan@msu.edu}
\orcid{0000-0002-7418-731X}

\author{Jiayu Zhou}
\affiliation{%
  \institution{Department of Computer Science and Engineering, Michigan State University}
  \streetaddress{428 S Shaw Ln Rm 3115, East Lansing, MI 48824}
  \city{East Lansing}
  \country{USA}}
\email{zhou@cse.msu.edu}
\orcid{0000-0003-4336-6777}

\author{Steve Drew}
\authornotemark[1]
\email{steve.drew@ucalgary.ca}
\orcid{0000-0003-4527-2635}
\affiliation{%
  \institution{Department of Electrical and Software Engineering, University of Calgary}
  \streetaddress{2500 University Drive NW}
  \city{Calgary}
  \state{Alberta}
  \country{Canada}
  \postcode{T2N 1N4}
}

\renewcommand{\shortauthors}{Wu and Drew, et al.}

\begin{abstract}
  The ultra-low latency requirements of 5G/6G applications and privacy constraints call for distributed machine learning systems to be deployed at the edge. With its simple yet effective approach, federated learning (FL) is a natural solution for massive user-owned devices in edge computing with distributed and private training data. 
  %
  %
  FL methods based on FedAvg typically follow a naive star topology, ignoring the heterogeneity and hierarchy of the volatile edge computing architectures and topologies in reality. 
  Several other network topologies exist and can address the limitations and bottlenecks of the star topology. This motivates us to survey network topology-related FL solutions.
  %
  %
  In this paper, we conduct a comprehensive survey of the existing FL works focusing on network topologies. 
  %
  %
  After a brief overview of FL and edge computing networks, we discuss various edge network topologies as well as their advantages and disadvantages. Lastly, we discuss the remaining challenges and future works for applying FL to topology-specific edge networks.
\end{abstract}

\ccsdesc[500]{General and reference~Surveys and overviews}
\ccsdesc[500]{Computing methodologies~Distributed computing methodologies}
\ccsdesc[500]{Networks~Network architectures}
\ccsdesc[500]{Computing methodologies~Machine learning}
\ccsdesc[500]{Computer systems organization~Distributed architectures}
\keywords{Topology-aware federated learning, star topology, tree topology, decentralized topology, hybrid topology, blockchain topology}

\maketitle

\section{Introduction}

Edge computing has been widely deployed in recent years as a strategy to reduce costly data transfer by bringing computation closer to data sources than conventional cloud computing. 


Both academia and industry have seen a surge in research relating to edge computing  \cite{shi2016edge, satyanarayanan2017emergence, liu2019survey}. Specifically, in the field of Industrial Internet of Things (IIoT)~\cite{pan2017future}, connected autonomous vehicles (CAVs) \cite{liu2019edge}, augmented reality (AR) \cite{al2017energy}, wearable technologies~\cite{chen2017empirical}, and hybrid architectures and systems combining cloud and edge \cite{wang2017cloud, xu2019computation, wang2020architectural, wang2021eihdp}. Edge computing is prevalent in agriculture, energy, manufacturing, telecommunications, and many other domains. It creates a tremendous amount of data at the edge with heterogeneous data distribution patterns. Distributed data of this scale and variety has created a persistent demand for machine learning to drive decision-making processes at the edge of the network.


Edge computing enables data and computational decentralization. Decentralized devices can collaboratively perform machine learning tasks together forming a cohesive network of nodes \cite{wang2019edges, mao2017survey}. We see the urgency and prospects that distributed learning systems will play a vital role as the bread and butter for edge-based decision-making applications \cite{yang2019federated, khokhar2022review}. Back in 2016, the EU imposed restrictions under the General Data Protection Regulation (GDPR) \cite{voigt2017eu}, which is a regulation in EU law regarding data protection and privacy. Commercial companies in the EU are forbidden from collecting, dealing with, or exchanging user information without their consent. China and the United States are implementing similar legislation as well \cite{yang2019federated}. As a result, federated learning (FL) has emerged as an optimal solution for edge computing without violating privacy legislation. Several companies have shown interest in FL applications, including Amazon \cite{Ding2022}, Google \cite{hard2018federated}, Webank \cite{he2019central}, IBM \footnote{https://github.com/IBM/FedMA} and ByteDance \footnote{https://github.com/bytedance/fedlearner}.

More recently, innovative approaches using foundation models \cite{zhang2023federated}, Vision Transformers (VIT) \cite{zhu2023xtab}, and pre-trained models \cite{hu2023federated, nguyen2022begin} are gaining huge attention. 
%
%
Despite its wide applications, critical challenges of edge computing~\cite{liu2019edge, qiu2020edge} remain in latency, communication costs, service availability, privacy, and fairness, especially for machine learning tasks at the edge. 
The low latency limits of communication to the cloud and data privacy requirements have inevitably grounded these distributed machine learning tasks: \textit{the data should never leave the edge}. 
The deployment of edge-based applications demands learning tasks to run at edge infrastructure instead of the remote cloud~\cite{guo2022distributed, wang2019edge}. 
The ubiquitous applications of edge computing in multiple industries \cite{zhang2020fenghuolun, kumar2021pefl} justify the strong motivation for the research of distributed machine learning in edge computing.

The topology of the edge network can sometimes be largely overlooked. In this survey, the network topology can be treated both as a challenge and a solution. As a challenge, specific topologies impose certain restraints like extra layers of communication and network structures. Whereas a solution in edge computing, topologies offer new ways to address different bottlenecks such as communication overhead, over-dependent to the central server, etc. Multiple topology structures exist in current FL works, and each topology brings its benefits and challenges. For example, ring topology \cite{lee2020tornadoaggregate} is utilized to enhance scalability and accommodate diverse client activities, thereby eliminating the need for a central server. \citet{hosseinalipour2020federated} propose fog learning, a paradigm that intelligently distributes ML model training across nodes, from edge devices to cloud servers. It enhances FL along three major dimensions: network, heterogeneity, and proximity.


\subsection{Scope and Contribution}
In this survey, we study the various network topology structures that exist in FL. 
In Table \ref{tab:past_surveys}, we compare our survey with existing surveys discussing network topology or FL. 
There is a significant gap in the number of surveys conducted on network topology at present. Since the introduction of FL in 2016, we have seen a huge increase in FL-related papers. Many Existing surveys look to treat network topology as a system limitation or challenge, while some papers propose to use new network topology to improve communication or computing efficiency. 
Several comprehensive surveys have extensively covered the general concepts, architectures, and applications in FL \cite{yang2019federated, kairouz2021advances, li2020federated, lim2020federated}.
More recently, many studies tend to concentrate on specialized areas within the FL domain, possibly due to the rapid rate at which FL works are being published. Some of the recent work includes FL in the health domain \cite{nguyen2022federated}, FL for the internet of things \cite{nguyen2021federated}, and blockchain-empowered FL \cite{zhu2023blockchain}. However, no existing surveys have discussed or organized research works of FL in edge computing from the network topology perspective. This gap leaves us with a vast open area to discuss FL from a new perspective.
Our study examines surveys from a wide range of dates of publication that discussed FL or network topology designs. We found that these two topics were never discussed together, even though FL comprises many different network topologies. This motivated us to 
showed them here in Table \ref{tab:past_surveys} about their difference to us.
\begin{table}[t]
    \centering
    \caption{Existing Surveys That Discuss FL Or Network Topology Design}
    \label{tab:past_surveys}
    \begin{tabular}{lcccc}
        \toprule
        Survey                        & Year & Focus                                     & Topology   & FL\\
        \midrule
        \citet{rajaraman2002topology} & 2002 & Topology and routing in Ad-hoc Network    & \checkmark & $\times$ \\
        \citet{li2013survey}          & 2006 & Overview of topology control techniques   & \checkmark & $\times$ \\
        \citet{donnet2007internet}    & 2007 & Measurements of network topology          & \checkmark & $\times$ \\
        \citet{lim2020federated}      & 2020 & FL in Mobile Edge Networks                & $\times$   & \checkmark \\
        \citet{kairouz2021advances}   & 2021 & FL Advances and Open Problems             & $\times$   & \checkmark \\
        \citet{nguyen2022federated}   & 2022 & FL for Smart Healthcare domains           & $\times$   & \checkmark \\
        \citet{nguyen2021federated}   & 2023 & FL Applications for IoT networks          & $\times$   & \checkmark \\
        \citet{zhu2023blockchain}     & 2023 & Blockchain-empowered FL                   & $\times$   & \checkmark \\
        \midrule
        Ours                          & 2024 & Edge Network Topology for FL               & \checkmark & \checkmark \\
        \bottomrule
    \end{tabular}
\end{table}
To our knowledge, no existing surveys have reviewed unique FL works from the edge network topology perspective and have promoted the development of various topology structures in FL. Compared with previous surveys, this paper's main contributions are: 
\begin{enumerate}
    \item Our survey introduces a novel perspective by employing edge network topologies (the network's structure) to categorize unique FL works.
    \item We provide a comprehensive classification of FL into four major topologies, including star topologies, mesh topologies, hybrid topologies, and less common network topologies, which provide clarifications for future research.
    \item We follow a systematic review approach using PRISMA \cite{moher2010preferred} for the paper selection process.
    \item We present the design, baselines, and benchmarks and then thoroughly review the key findings of some highlighted work.
    \item We outline promising research directions and challenges for the future development of topology-aware FLs.
\end{enumerate}
The rest of the paper is organized as follows. In Section \ref{sec:RM}, we explain our research methodology. In Section \ref{sec:3over}, we introduce an overview of FL in edge computing. In Section \ref{sec:type}, we propose eight types of FL network topologies and summarize existing studies along each topology. In Section \ref{sec:chall}, we present some of the open Issues in edge FL Topology. In Section \ref{sec:chall}, we explain the limitations and synthesize a roadmap for future research. Last, we conclude our paper in Section 6.
\begin{figure}[t]
    \centering
    \includegraphics[width=0.68\linewidth]{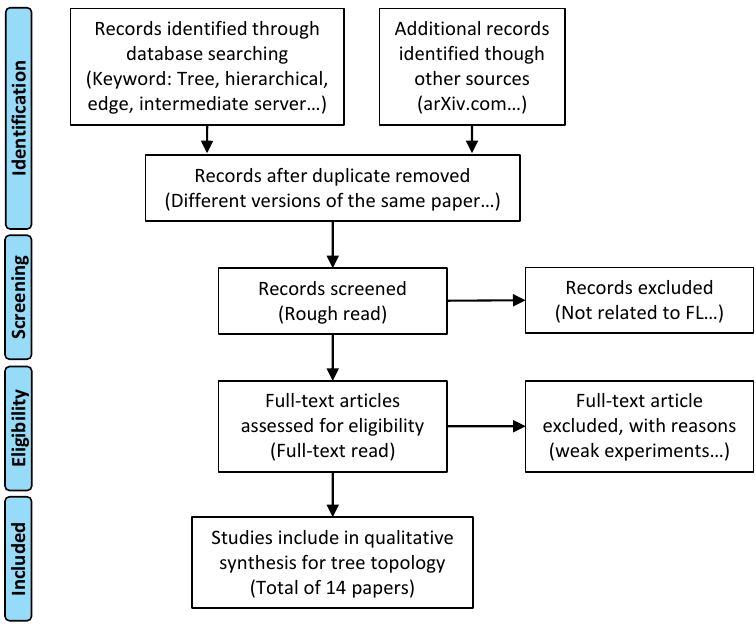}
    \caption{The PRISMA flow diagram with reasoning for tree topology.} 
    \Description[PRISMA flowchart outlining the stages of selection for this study]{The figure presents a flow diagram that outlines the systematic approach to selecting research papers for a review using the PRISMA. Initially, records are identified through database searches using specific keywords such as Tree, hierarchical, edge, and intermediate server. Additional records are sourced from other repositories like arXiv.com. Following identification, duplicates are removed, citing the example of different versions of the same paper being excluded. The next stage involves screening of records based on a rough read, leading to the exclusion of records not related to FL. The remaining records undergo a full-text review for eligibility, further filtering out articles with reasons such as weak experiments. The final stage includes studies selected for qualitative synthesis based on tree topology, noting 14 papers were included at the end.} 
    \label{fig:PRISMA}
\end{figure}

\section{Research Methodology}
\label{sec:RM}
\subsection{Research Goals Formulation}
We aim to provide an in-depth and systematic overview of all papers on FL that utilize one or more unique network topologies. Furthermore, we use the PRISMA \cite{moher2010preferred} search strategy to collect all the papers following a similar approach as \citet{pfitzner2021federated}. We show an example of searching for tree topology papers using the PRISMA flow diagram in Fig. \ref{fig:PRISMA}. Additionally, we aim to show evidence that different network topologies and FL can benefit from each other. We summarize our research goals into three points.
\begin{itemize}
    \item Identify existing edge network topology structures in the current FL literature.
    \item Examine the unique challenges and benefits different topologies bring.
    \item Provide readers with an overview of the baseline methods and datasets used in each paper.
\end{itemize}
\subsection{Search Strategy}
\label{sec:searchstra}
For our paper search strategy, we start by searching papers that contain the terms “\textit{federated learning} \textit{AND} \textit{topology}.” We submitted this search criteria to the digital scholarly databases. The three main scholarly databases we used are the digital library of the Association for Computing Machinery, the online portal of the Institute of Electrical and Electronics Engineers, and Google Scholar. This search strategy only returned a few papers and it was not helpful. Therefore, we modified our search strategy to treat every topology as its own branch of work and restarted the search process. The detailed search process is listed below: 
\begin{enumerate}
    \item Identity a topology structure to start the search process (star, tree, ... or mesh topology).
    \item Initalite the PRISMA \cite{moher2010preferred} search process.
    \item Group the specific topology paper into major topology or minor topology.
\end{enumerate}
\subsection{Inclusion and Exclusion Criteria}
\begin{figure}[htb]
    \centering
    \includegraphics[width=0.6\linewidth]{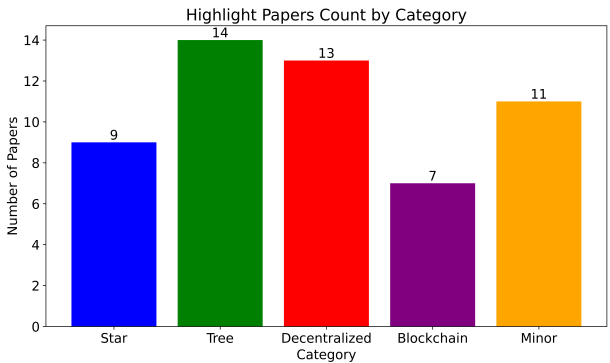}
    \caption{Number of papers for each network topology type.} 
    \Description[Bar Graph of Network Topology Types Based on Highlight Papers Count]{The Tree topology has the highest number of papers at 14, followed by the Decentralized topology with 13 papers. The Star topology has 9 papers, while the Blockchain topology has the fewest with 7 papers. The Minor topology includes 11 papers.} 
    \label{fig:papercount}
\end{figure}

Our survey aims to give readers a good understanding of FL, and the upsides and downsides of several network topology structures, so we have selected the following criteria for inclusion. After collecting the papers returned from the database search, we include papers that are:
\begin{itemize}
    \item Peer-reviewed (Identification phase)
    \item Presenting one or more unique topology structures (Identification phase).
    \item Using FL as the primary methodology (Screening phase).
    \item Implanting and comparing the proposed method with strong baseline methods (Eligibility phase).
\end{itemize}
However, the surveyed query terms return many irrelevant works to this review. Some papers may contain only one or two mentions of FL and cover completely unrelated topics. Our exclusion criteria are listed below:
\begin{itemize}
    \item Share different titles but are different versions of the same paper (Identification phase).
    \item FL is not involved at all (Screening phase).
    \item Experiments do not use known baseline methods (Eligibility phase).
    \item Application-centring or case study (Eligibility phase).
    \item Benchmark paper evaluating existing works (Eligibility phase).
\end{itemize}
We use our proposed strategy to search for 8 topologies in section \ref{sec:searchstra}. We organize the results into 5 major topology types shown in Fig. \ref{fig: FL topology}.
A total of 42 papers meet all selection criteria. We also illustrate the number of papers from each topology in Fig. \ref{fig:papercount}. 
%

\section{An overview of Federated Learning in Edge Computing}
\label{sec:3over}

\subsection{Background}
Recent years have embraced effervescent advances of Federated Learning (FL) algorithms in various applications, including IoT \cite{mills2019communication}, healthcare \cite{rieke2020future}, image processing \cite{khokhar2022review}, etc.

In the representative federated learning approach \textit{FedAvg}~\cite{mcmahan2017communication}, with the restrictions of GDPR \cite{voigt2017eu}, each mobile device learns a local model and updates the model to a central server periodically. A central server then aggregates local models using a simple yet effective method to produce a global model and distributes the global model to all Android devices for the next learning cycle.
The FedAvg algorithm improves over FedSGD, which uses parallel stochastic gradient descent (SGD). 
FedSGD selects a set of workers each round, and the selected workers compute the gradient using the global model parameters and their local data. 
Gradients from each worker are sent back to the server, which performs SGD using the combined gradients and the learning rate. The process is repeated until the model converges. 
Compared to the baseline algorithm FedSGD, FedAvg requires significantly fewer rounds of communication to converge.
As with FedSGD, FedAvg followed the general computation steps where the server sends the model parameters to each worker. 
Each worker computes the gradient using the received model parameters, its local data, and a given learning rate. 
FedAvg differs from FedSGD in that each worker repeats the training process multiple times before sending the updated model parameters back to the server.
FedAvg was developed with the intention of achieving the same level of efficacy with less communication to the server. While the overall computation task for each worker increased, there were fewer rounds of communication compared to FedSGD, resulting in a trade-off between computation and communication costs. 
In many FL scenarios, the edge clients generally have limited data residing locally\cite{mcmahan2017communication, bonawitz2019towards}. Even though deep models are commonly used, the computational expenses are often overshadowed by the communication costs. This is why FL with FedAvg algorithm \cite{mcmahan2017communication}, known for its communication efficiency, is particularly effective.

We categorize the FL algorithms we surveyed based on their emphasis on the types of challenges they tackled.

\subsubsection{Statistical and System Heterogeneity}
A significant amount of effort has been made to address the issue of \textit{user heterogeneity} in FL. 
Specifically, the heterogeneity is manifested in both \textit{statistically} and \textit{systematically}.

\begin{figure}[htb]
    \centering
    \includegraphics[width=0.48\linewidth]{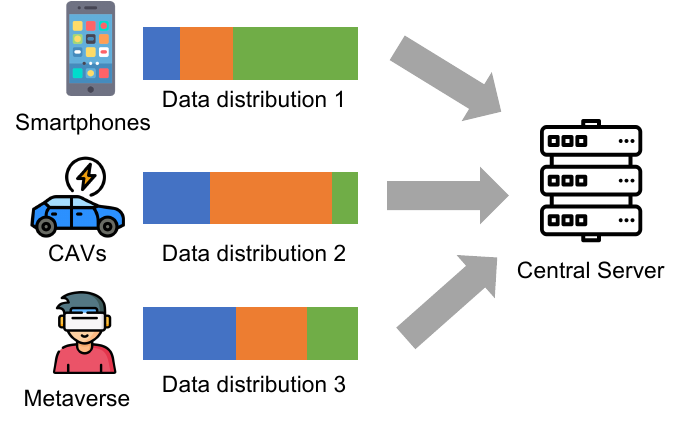}
    \includegraphics[width=0.48\linewidth]{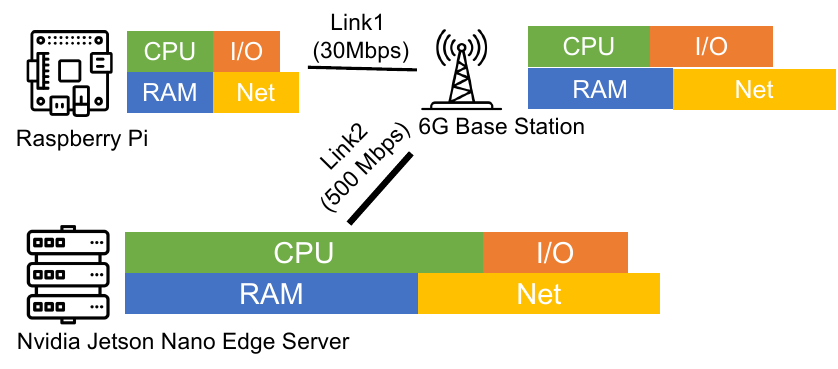}
    \caption{\textbf{Right: }An example showing the statistical heterogeneity among different types of clients in FL. Depending on how the clients generate data, the statistical distributions and patterns of data on each device can be very different. \textbf{Left: }A demonstration of system heterogeneity. Three different tiers of edge devices have distinct capabilities of computing, connected by links with different bandwidths.} 
    \Description[Representation of Data Distribution Across Devices and Servers]{Right figure: A central server connected with Smartphones, CAVs, and Metaverse devices with different data distribution. Left figure: Raspberry Pi device and Nvidia Jetson Nano Edge Server are connected with 6G Base Station.}
    \label{fig:data-heterogenity}
\end{figure}
On the one hand, \textit{statistical heterogeneity} in FL refers to the differences among user local data distributions as shown in Fig. \ref{fig:data-heterogenity}. 
Namely, the clients have datasets that are not independent and identically distributed (Non-IID). 
When the data is collected locally, such differences are likely induced by heterogeneous user behavior.
%
In the case of Non-IID data distributions, aggregation may lead to a biased global model with a sub-optimal generalization performance. 
This phenomenon is also known as client-drift \cite{karimireddy2020scaffold}, and it refers to the process by which global models are updated toward local optimal solutions as a result of heterogeneous data.

Towards addressing this client-drift issue, previous works including FedProx \cite{sahu2018convergence}, pFedMe \cite{t2020personalized}, and SCAFFOLD \cite{karimireddy2020scaffold}, have proposed constraining the local model parameters to prevent them from diverging far from the global model optimal. 
\textit{Personalized FL} is an alternative strategy for handling data heterogeneity. It permits different model parameters or even architectures to be adopted by local users. Besides diversified architectures, few-shot adaptations can also achieve personalization by fine-tuning a global model using local data~\cite{t2020personalized, li2021ditto}.

In the meantime, \textit{system heterogeneity} results from different user capacities in terms of computation, memory, bandwidth, etc. We show an example of system heterogeneity in Fig. \ref{fig:data-heterogenity}.
Adopting one unified model architecture for FL can be undesirable under such scenarios: an over-large global model might bring heavy workloads to small users which lack computation or transmission resources, while an over-small global model may under-perform in capturing complex feature representations for the learning tasks.
Therefore, an emerging group of algorithms is pursuing FL frameworks that support heterogeneous user model architectures \cite{zhang2018joint, ma2021fedsa, zhang2016energy, hosseinalipour2020federated, wang2021device, zehtabi2022decentralized, sahu2018convergence}.

\subsubsection{Privacy}  
Although FL allows decentralized devices to participate in machine learning without directly exchanging data, there are still potential privacy concerns. Furthermore, adversaries may be able to deduce some original data from the parameters of a model. 

High-level FL privacy threats include inference attacks and communication bottlenecks. Secure multi-party computation, differential privacy, verifyNet, and adversarial training are effective techniques for preserving privacy in FL \cite{mothukuri2021survey}. 

\subsubsection{Convergence Guarantee}  
There have been extensive studies about the theoretical convergence properties of FL algorithms under different problem settings. Pioneer efforts along this line, such as \cite{mcmahan2017communication, li2019convergence, qu2020unified},  have analyzed the convergence speedup of FL algorithms and derived a desirable conclusion that linear speedup can be achieved for FedAvg, which is the most representative FL algorithm, with commonly adopted assumptions for analysis.
%
%

\subsubsection{Communication efficiency}  
To improve communication efficiency, one popular approach is to either reduce the number of communication rounds or to require fewer data to be transmitted per communication round \cite{konevcny2016federated}. 
Depending on the infrastructure of computing, by selecting an appropriate topology design, communication efficiency can also be optimized.
Generally, the star topology network design ensures the least amount of communication with the central server since all devices are directly connected to it. 
In tree topologies, intermediate edge servers are usually involved, and devices can benefit from fast and efficient communication with edge devices at a low cost.
For fully meshed topologies, communication usually takes place in a P2P or D2D manner, and direct communication between the devices is generally quite efficient.
Furthermore, hybrid topologies are emerging, which combine the common topology with the strengths of each to produce a more dynamic system.

The aforementioned challenges in FL can be tackled in parallel. For instance, some personalized FL algorithms \cite{t2020personalized, li2021ditto},  which share only partial model parameters, can tackle user heterogeneity while achieving high communication efficiency.

\begin{figure*}[tb]
    \centering
    \includegraphics[width=0.53\linewidth]{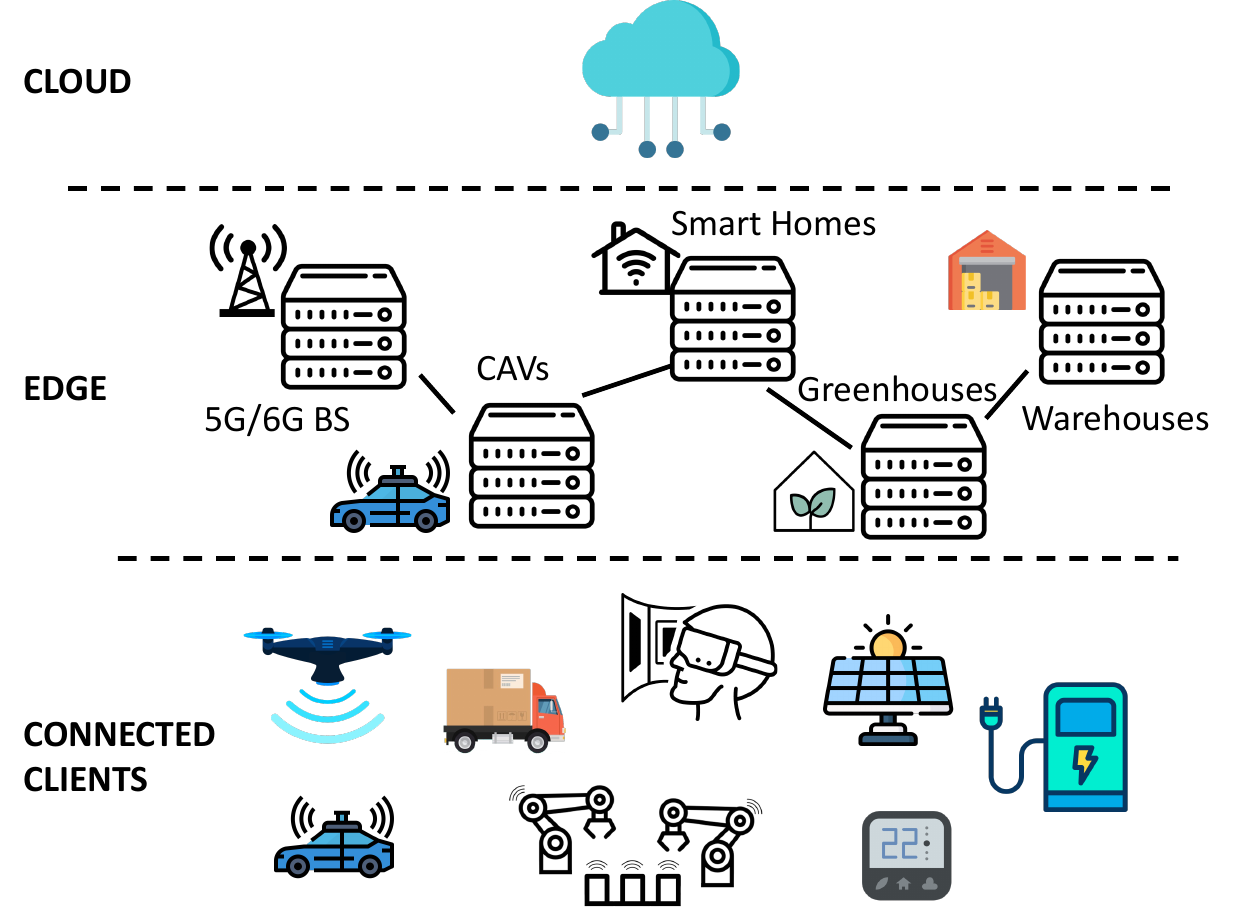}
    \caption{Application scenarios of federated learning in mobile edge computing.}
    \Description[Various application scenarios of FL in mobile edge computing, categorizing different entities into connected clients, edge infrastructure, and the cloud.]{Application scenarios of FL in mobile edge computing include edge 5G/6G BS, CAVs, Smart Homes, Greenhouses, and Warehouses. Connected devices such as drones, trucks, self-driving cars, manufacturing, solar panels, temperature controllers, and chargers.}
    \label{fig:fl-mec-scenarios}
\end{figure*}

\subsection{FL Characteristics Specific to Edge Computing}

Unlike the typical configurations of FL, which follow a naive star topology, practical edge computing demonstrates unique characteristics in architectures and network configurations, which could deeply impact the design and implementation of the effective deployment of FL algorithms. We list the key features of edge computing frameworks below.

\subsubsection{Heterogeneity, energy efficiency, and task offloading}  
A large portion of edge networks consists of user devices. These devices include highly embedded devices such as wearable glasses and watches, as well as powerful personal servers \cite{shi2016promise, yu2017survey, siriwardhana2021survey}. 
These devices are mostly still powered through batteries, making it necessary to consider \textit{energy-efficient} protocol and algorithm design \cite{jiang2020energy, sun2019joint}. 
The heterogeneous devices also introduce a huge variance of computational capacities, leading to a natural research direction of task offloading \cite{chen2018task, tran2018joint}.

The heterogeneity, energy efficiency, and task offloading play substantial roles in formulating the topologies of the edge networks \cite{ning2020heterogeneous}. 
To elaborate, energy consumption considerations prohibit a star-topology in a large-scale edge network because the central edge server would be overwhelmed due to its capability limit \cite{ma2021fedsa}. 
Offloading FL tasks from less capable edge devices to more powerful edge devices is a viable and increasingly researched approach in FL and edge computing \cite{chen2018task, tran2018joint, sun2019joint}.
The offloading schemes are typically accompanied by their corresponding topology best practices \cite{zhang2016energy, zhang2018joint}.

\subsubsection{Hierarchy and clustering}  
The nature of edge computing and 5G/6G communications has led to hierarchical networks where a base station covers the data transmission in small areas of wireless edge devices \cite{kiani2017toward, cong2020survey}. 
The partial coverage results in multiple base stations deployed at the edge networks. The base stations can forward the data to the central server in a three-tier network hierarchy. When the number of base stations is large enough, there can be even more than three tiers for the data to move up along the hierarchy. The multi-fold hierarchy creates the prerequisites for configurable clustering and aggregation, making space for creativity over the hierarchical edge networks.

Hierarchical networks at the edge are often another product of the heterogeneous edge devices. Separations of capabilities have evolved into separations of hierarchies. Edge devices with lower capabilities can be dedicated to collecting sensing data and uploading it to their edge servers, whereas the edge server can be used for training local models and receiving updated global models. To minimize the exposure of the models to non-server parties, the edge servers can use different topology patterns from the central server to communicate with each other to maximize efficiency and privacy. In multi-tier edge networks, dynamic topologies can be applied based on internal and external factors for optimal learning performance. 

\subsubsection{Availability and mobility}  
Compared to cloud data centers, edge servers have less redundancy and less reliability due to space, power, and budget. Mobile Edge devices, such as CAVs and unmanned aerial vehicles (UAVs), have even lower availability because of their mobility. The moving edge device may enter and exit the boundaries of an edge network and switch between different clusters of an edge network, leading to interruptions of task processing and computations. Fig. \ref{fig:fl-mec-scenarios} shows some application scenarios of mobile edge computing in FL.

The mobile and volatile edge devices of CAVs and UAVs are pushing dynamic-format topology, where at any epoch of the system, there can be the addition or removal of edge devices and edge servers. 

\subsection{FL Challenges and Solutions in Edge Network Topologies}

The unique characteristics of edge computing and edge networks are posing fundamental challenges to performing reliable and efficient federated learning and applying feasible distributed learning systems at the edge. Many of those challenges can be resolved or mitigated by topology designs. In the following sections, we list some of the major challenges in FL and their corresponding solutions using a different network topology structure. 

\subsubsection{Scattered data across organizations}  
As its name describes, FL may require data from independent organizations to be federated. In this scenario, there are stricter data-sharing policies without directly sharing any data or the intermediate local models. 
For example, federated transfer learning (FTL) \cite{liu2020secure} can unite those organizations and leverage their data without violating privacy protection regulations. Compared with the vanilla FedAvg, FTL allows learning from the entire dataset, rather than only those samples with common features.

\subsubsection{High communication costs}  
The original FL requires each device to directly communicate with the central server for upstream model aggregation and downstream model update. In the context of edge computing, direct communication to the central server is expensive for some edge devices and may cause high latency. The hierarchical edge computing topology can pool and aggregate the local updates from devices and hence reduce the communication costs to the cloud.

\subsubsection{Privacy concerns and trust issues}  
While federated learning keeps the storage of training data to the device, it still does not eliminate the risk of exposing sensitive information through repeated aggregated local model uploads to central servers. 
When a threat model considers the privacy concern in central aggregation servers, a network topology with decentralized model aggregation methods will help mitigate or eliminate the risk.
The rationale behind this is that all relaying edge servers in the topology will aggregate part of the information. So one compromised central server will not be able to see all the fine-granularity model updates from all clients, and therefore largely reduces the differential information repeatedly exposed to the server. 

\subsubsection{Imbalanced data distribution}  
The nature of heterogeneous edge devices and networks in edge computing environments has led to significantly imbalanced data distribution and intensity based on the type of applications and devices. For example, an augmented reality (AR) application may generate a large burst of data over a short period when a user is actively using the application. In comparison, a temperature monitoring application may only generate a small amount of data for temperature records. However the data is produced constantly and periodically. By utilizing the tree network topology, methods like Astraea \cite{duan2020self} add mediators between the FL server and the clients to resolve imbalanced data problems.

\subsection{Categorization of Topology-aware FL in Edge Computing}

With the recent advancements in deep learning and increasing research interests in FL, a growing number of studies have expanded the horizon of FL applications. Numerous studies have reviewed existing FL areas \cite{li2021survey, mothukuri2021survey, lim2020federated, khan2021federated}. However, due to the broad applications and the nature of FL, there is no standard to summarize existing Topology-aware FL studies systematically. Many existing FL studies focus on specific characteristics of FL and attempt to categorize it accordingly. Several of these methods are summarized in the following section. 

\subsubsection{Based on Data Partition:
Horizontal FL (HFL), vertical FL (VFL), and federated transfer learning (FTL)}  
FL can be categorized into horizontal FL (HFL), vertical FL (VFL), and federated transfer learning (FTL) based on data partition in the feature and sample spaces \cite{yang2019federated}. 
Horizontal FL (HFL) represents a typical FL setting where the set of features of all participating clients are the same, making it easy for implementation and training. 
In most cases, studies treat horizontal FL as the default structure and may not even mention the term ``horizontal''. 
For example, the first implementation of FL by Google \cite{mcmahan2017communication} is an example of HFL where the feature space of all participating devices is the same.

On the other hand, Vertical FL (VFL) \cite{wei2022vertical} is catered specifically toward vertically partitioned data, where clients in VFL have different feature spaces. For example, hospitals and other healthcare facilities may have data about the same patient but different types of health information. Fusing multiple types of information from the same set of samples or overlapping samples in different institutions belongs to the VFL setting.  

FTL \cite{liu2020secure} was initially designed for scenarios where participants in FL have heterogeneous data in both feature space and sample space. 
In this setting, both HFL and VFL are unable to train efficiently in heterogeneous settings. FTL is considered the ideal solution at the time. FTL leverages the whole sample and feature space with transfer learning, where two neural networks serve as feature transformation functions that can project the source features of the two networks into a common feature subspace, allowing the knowledge to be transferred between the two parties. 

\subsubsection{Based on Model Update Protocols: Synchronous, Asynchronous, and Semi-synchronous FL}  
FL can be separated into synchronous, asynchronous, and semi-synchronous by communication protocols \cite{stripelis2021semi}. 
For synchronous FL \cite{gupta2018distributed,vepakomma2018split,liu2019communication}, each learner performs a set round of local training. After every learner has finished their assigned training, they share their local models with the centralized server and then receive a new community model, and the process continues. Synchronous FL may result in the underutilization of a large number of learners and slower convergence, as others must wait for the slowest device to complete the training. 
With asynchronous FL \cite{stripelis2021semi, xie2019asynchronous}, there are no synchronization points. Instead, learners request community updates from the centralized server when their local training has been completed. As fast learners complete more rounds of training, they require more community updates, which would increase communication costs and lower the generalization of the global model. 
A semi-synchronous FL framework called FedRec \cite{stripelis2021semi} was proposed that allowed learners to continuously train on their local dataset up to a specific synchronization point where the current local models of all learners are mixed to form the community model.

\subsubsection{Based on Data Distribution: non-IID and IID data FL}  
One of the major statistical challenges surrounding FL in the early stage is when training data is non-IID \cite{zhao2018federated}. The consistent performance of FL relies heavily on IID data distribution on the local clients. 
However, in most real-life cases, local data are likely non-IID, which significantly decreases the performance of existing FL techniques if not catered specifically to non-IID data. Therefore, existing FL studies can be categorized as FL with non-IID data or FL with IID data. 

\subsubsection{Based on Scale of Federation: Cross-silo and Cross-device FL}  
Based on the scale of the federation, FL studies can be divided into cross-silo and cross-device FL \cite{kairouz2021advances}. Cross-silo FL focus on coordinating a small amount of large data centers like hospitals or banks. On the other hand, Cross-device FL has relatively large amounts of devices and small amounts of data in each device. The key differences between the two are the number of participating parties and the amount of data stored in each participating party in FL.

\subsubsection{Based on Global Model: Centralized and Decentralized FL}  
The most straightforward method for implementing and managing FL is to connect all participating devices through a central server. For centralized FL, the central server is either used to compute a global model or to coordinate local devices \cite{ni2022star}. Having a central server, however, may contradict the aim of decentralization in FL. For fully decentralized FL, there is no overarching central server at the top, and devices are connected in a D2D or P2P manner.

\section{Types of FL Network Topology}
\label{sec:type}
To inspire future research, our work summarizes state-of-the-art FL studies from the perspective of network topology, as opposed to existing FL reviews that focus on particular features, such as data partitioning, communication architecture, or communication protocols.

\begin{figure*}[htb]
    \centering
    \includegraphics[width=0.95\linewidth]{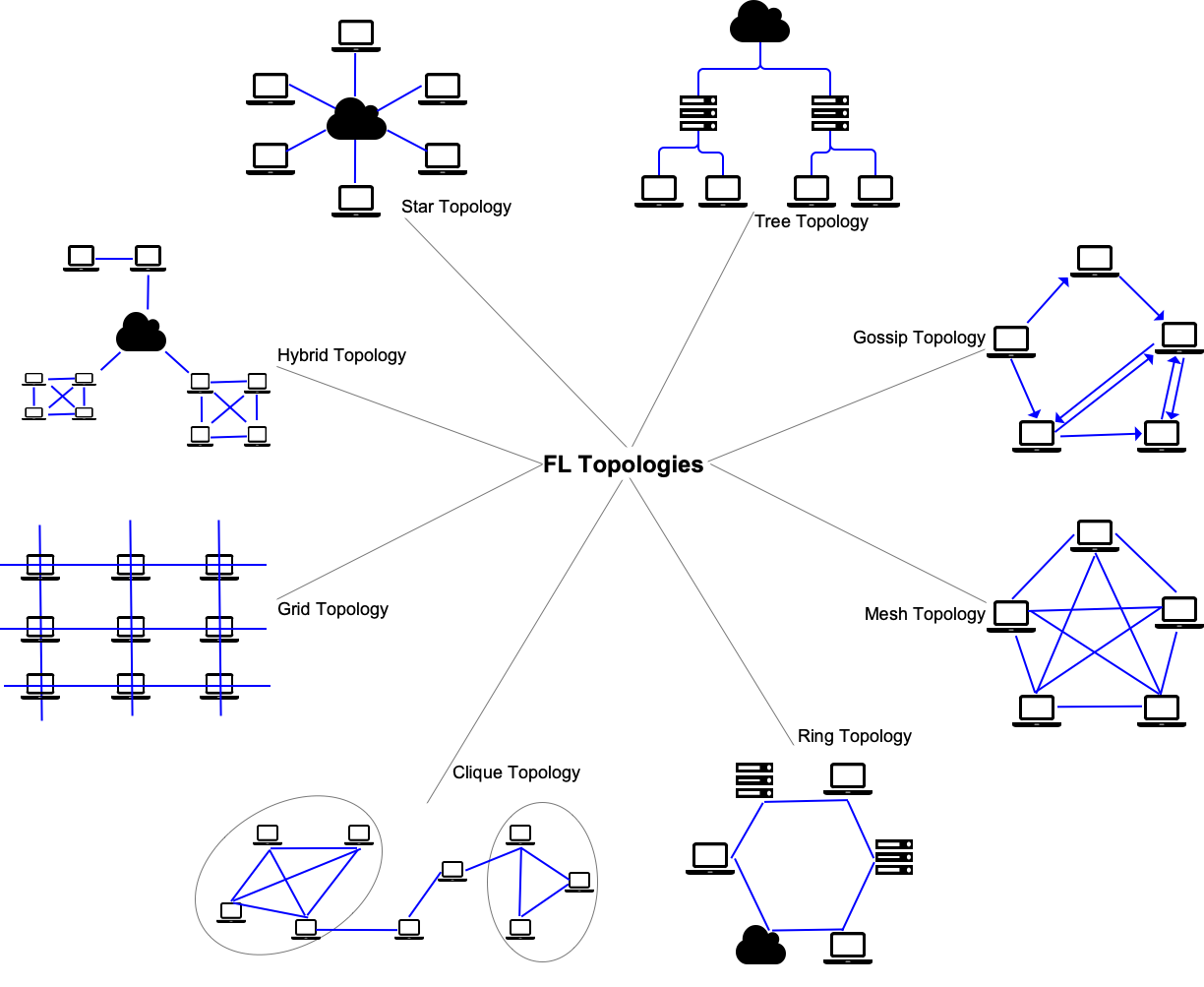}
    \caption{Overview of FL topology types.}
    \Description[Diagrams of various FL topologies]{Star Topology features a central node connected to all peripheral nodes. Tree Topology shows a hierarchical structure with one top-level central node connected to intermediary nodes, which are connected to peripheral nodes. Gossip Topology displays a non-hierarchical network where each node is connected to several others. Mesh Topology illustrates an all-to-all connection where every node is interconnected with every other node in the network. Ring Topology presents nodes connected circularly. Grid Topology depicts nodes arranged in a grid pattern, connected to adjacent nodes horizontally and vertically. Hybrid Topology combines elements of other topologies.}
    \label{fig: FL topology}
\end{figure*}

%
In the case of FL, the network topology represents how edge devices communicate with each other and eventually to a centralized server \cite{zhang2021survey}. FL can benefit from the topologies of the networks to increase communication efficiency by performing partial and tiered model aggregations \cite{chai2020fedat,chai2020tifl}, enhancing privacy by avoiding transmitting local models directly to a centralized server \cite{zhou2021two,zhou2020privacy}, and improving scalability with horizontally replicable network structures \cite{charles2021large,briggs2020federated}. 
The major types of FL network topology essentially come down to either centralized (e.g., star topologies), decentralized (e.g., mesh topologies), or hybrid topologies, which consist of two or more traditional topology designs.
Other less common network topologies, like ring topologies, will also be covered in this section. Fig. \ref{fig: FL topology} shows the overview of the mentioned FL topology.

\subsection{Star Topology}
The original use case of FL was to train machine learning algorithms across multiple devices in different locations. The key concept is to enable ML without centralizing or directly exchanging private user data. However, most FL implementations still require the presence of a centralized server. 
The most common network topology used in FL, including the original FL work \cite{mcmahan2017communication}, adopted centralized aggregation and distribution architecture, also known as ``star topology''. As a result, the graph of the server-client architecture resembles a star. Numerous FL research and algorithms are based on the assumption of a star topology \cite{gupta2018distributed, vepakomma2018split, ni2022star}. 
While being the most straightforward approach, a star network topology suffers from issues like high communication costs, privacy leak concerns to the central server, and security concerns \cite{fereidooni2021safelearn}. Some studies posed solutions to address these issues \cite{ni2022star}. However, the star-topology-based solutions are not always the optimal network topology design for all FL systems. It is worth questioning if the star architecture is the network topology that best fits all scenarios.

There are a substantial amount of studies in FL using the default star network topology \cite{yu2017survey, liu2019survey, yang2019federated, kairouz2021advances, li2020federated, lim2020federated}. Most of those studies do not focus on the aspect of network topology or edge computing. In this section, we select various FL works that focus on optimizing the topology, communication cost, and edge computing while still using the traditional star topology structure. In Table \ref{table:highlights_star}, We highlight some of the works using star topology.

\begin{table}[htb]
\footnotesize
\centering
\caption{Highlighted works - Star Topology}
\label{table:highlights_star}
\renewcommand{\arraystretch}{1.5}
\begin{tabularx}{.99\textwidth}{|l|p{4.8cm}|X|}
\hline
\textbf{FL Type} & \textbf{Baselines \& Benchmarks} & \textbf{Key Findings } \\ \hline

\multirow{5.5}{46pt}{Synchronous}
 & FedAvg and Large Scale SGD with MNIST, CIFAR-10, CIFAR-100, and ILSVRC 2012 & Computation and communication bandwidth were significantly decreased \cite{gupta2018distributed, vepakomma2018split} \\ \cline{2-3}
 & FedSGD, FedBCD-p, and FedBCD-s with MIMIC-III, MNIST, and NUS-WIDE & The models performed as well as the centralized model. Communication costs were significantly reduced \cite{liu2019communication} \\ \cline{2-3}
 & Noise-Free FL, Conventional RIS, Random STAR-RIS, Equal Power Allocation with MNIST, CIFAR-10 under IID \& non-IID & STAR-RIS used both NOMA and AirFL framework to address the spectrum scarcity and heterogeneous services issues \cite{ni2022star} \\ \cline{1-3}

\multirow{3.5}{46pt}{Asynchronous/ Semi-Synchronous}
 & FedAvg and single-thread SGD with CIFAR-10 and WikiText-2 & FedAsync was generally insensitive to hyperparameters, had fast convergence and staleness tolerance \cite{xie2019asynchronous} \\ \cline{2-3}
 & FedAvg, FedAsync, and FedRec with Cifar-10 and Cifar-100 & Faster generalization \& learning convergence, better utilization of available resources and accuracy \cite{stripelis2021semi} \\ \cline{1-3}

\multirow{6.5}{46pt}{Personalized}
 & eFD(Extended Federated Dropout) and Federated Dropout(FD) using CIFAR10, FEMNIST, and Shakespeare & Able to extract submodels of varying FLOPs and sizes without the retraining; flexibility across different environment setups \cite{horvath2021fjord} \\ \cline{2-3}
 & pFedMe, Ditto, FedAlt, and FedSim with StackOverflow, EMNIST, GLDv2, and LibriSpeech &  Proposed partial model personalization can obtain most benefit of full model personalization; provided convergence guarantee \cite{pillutla2022federated} \\ \cline{2-3}
 & FedAvg, pFedMe, Ditto, FedEM, FedRep, FedMask, and HeteroFL with EMNIST, FEMNIST, CIFAR10, and CIFAR100 & Significantly improves performance; thorough theoretical analysis; extensive experiments are conducted show superior effectiveness, efficiency, and robustness \cite{chen2023efficient} \\ \cline{1-3}
 
\hline
\end{tabularx}
\end{table}


A distributed learning method called splitNN was proposed \cite{gupta2018distributed, vepakomma2018split} to facilitate collaborations of health entities without sharing the raw health data. In a star topology, all subsequent nodes are connected to the master node. Data does not have to be shared directly with the master node. By using a single supercomputing resource, a star topology network can provide training with access to a significantly larger amount of data from multiple sources. Alice(s) represent data entities in the deep neural network, and Bob represents one supercomputing resource that corresponds to the role of nodes and a central server. While all the single data entities (Alices) are connected to the supercomputing resource (Bob), no raw data are shared between each other. Techniques will include encoding data into a different space and transmitting it to train a deep neural network. Experimental results were obtained on the MNIST, CIFAR-10, and ILSVRC (ImageNet) 2012 datasets and showed similar performance to other neural networks trained on a single machine. As compared with classic single-agent deep learning models, this technique significantly reduces client-side computational costs.
Although federated learning was available at the time, the authors argued that there had been no proper non-vanilla settings with vertically partitioned data and without labeling, with distributed semi-supervised learning and distributed multi-task learning.

An algorithm named Federated Stochastic Block Coordinate Descent (FedBCD) \cite{liu2019communication} was proposed boasting multiple local updates before communications to the central server.
Through theoretical analysis, the authors found that the algorithm needed $O(\sqrt{T})$ iterations for T iterations and achieved $O(1/\sqrt{T})$ accuracy.

\citet{ni2022star} proposed a new FL framework called STAR-RIS which integrates nonorthogonal multiple access (NOMA) and over-the-air federated learning (AirFL). STAR-RIS used NOMA and AirFL frameworks to address the spectrum scarcity and heterogeneous services issues in FL. This work follows the classical star topology where all client needs to update in a synchronized fashion and connect to the server. The proposed STAR-RIS used a novel approach that utilizes simultaneous transmitting and reflecting reconfigurable intelligent surface that boosts performance compared with other methods.
STAR-RIS addressed issues specific to the integration of communication and learning technologies for the 6G network. STAR-RIS provided a closed-form expression for the convergence upper bound, which gives a strong theoretical guarantee.

\subsubsection{Asynchronous FL topologies}  
\citet{stripelis2021semi} identified the issue that in heterogeneous environments, classic FL approaches exhibited poor performance.
Synchronous FL protocols were communication efficient but had slow learning convergence, while asynchronous FL protocols had faster convergence but higher communication costs. For synchronous FL, the original FedAvg algorithm serves as a great example: after each participant device trains for a fixed number of epochs, the system will wait until all the devices complete their training and then compute the community models. 
%
The approach is in no way efficient, but it limits communication to a fixed amount because all devices will have the same number of communication rounds.
In particular, in the case of fast and slow workers, the fast devices will have a long idle time to wait for the slow devices. For asynchronous FL, FedAsync \cite{xie2019asynchronous} provides a thorough analysis of the subject. asynchronous FL is the complete opposite of synchronous FL. As the previous protocol minimizes communication costs, asynchronous protocols seek to utilize all participant devices to their fullest capability, which means once a device finishes assigned training, it can request a community update and continue training. However, this approach significantly increases network communication costs for fast devices. Semi-synchronous \cite{stripelis2021semi} FL seeks to combine the benefits of both protocols by setting up a synchronization point for all devices. Allowing the fast devices to complete more rounds of training and prevent excessive communication along the way.

\subsubsection{Personalized star topology}  
There has been great attention to personalized FL in recent studies, mainly to increase the fairness and robustness of FL \cite{tan2022towards}. Most personalized FL \cite{li2021ditto, chen2023efficient, tan2022towards, horvath2021fjord} follows the traditional star topology. One interesting aspect of personalized FL is that personalized local clients may require fewer models to be transmitted over the network, one approach is to partially upload and download the global models from the server \cite{pillutla2022federated}. Another approach is to have a dynamically adapting model size based on the heterogeneous data distributions or resource constraints \cite{chen2023efficient, horvath2021fjord}. From the topology perspective, personalized FL brings some unique opportunities for further optimizing communication with various-sized local models.

\citet{pillutla2022federated} explored the idea of training partially personalized models, each local model has some shared and personal parameters. The authors experimented with both the simultaneous update and alternating update approaches. In addition, there is another personalized FL known as pFedme \cite{t2020personalized}, which employs Moreau envelopes as a way of regularizing loss functions. pFedMe follows the same structure as the conventional Fedavg algorithm with an additional parameter used for the global model update. Specifically, each client must solve to obtain their personalized mode, which is used for local updates. The server uniformly samples a subset of clients and the local model is sent to the server. \citet{horvath2021fjord} proposed Fjord which dynamically adapts the model size by with Ordered Dropout. By using this importance-based pruning approach, Fjord can create nested submodels from a main model and enable partial training only on the submodels. Fjord shows strong scalability and adaptability compared with baseline methods. \citet{chen2023efficient} take a further step on personalized FL by optimizing both clients' local data distribution and hardware resources using adaptive gated weights. The proposed pFedGate \cite{chen2023efficient} can generate personalized sparse models while also considering the resource limitation of the local device. Combining both model compression and personalization approaches, pFedGate achieves superior global and individual accuracy and efficiency compared to existing methods.

\subsubsection{Cohorts \& Secure aggregation}  
Charles et al. \cite{charles2021large} studied how the number of clients sampled at each round affected the learning model. Challenges were encountered while using large cohorts in FL. Particularly, the data heterogeneity caused the misalignment between the server model $x$ and the client's loss $f_k$. With a threshold for ``catastrophic training failure'' defined, the authors revealed that the failure rate increased from 0\% to 80\% when the cohort size expanded from 10 to 800. While the star topology remained the same, improved methods were proposed, including dynamic cohort sizes \cite{smith2017don}, scaling the learning rate \cite{goyal2017accurate, krizhevsky2014one}.

Secure aggregation protocols with poly-logarithmic communication and computation complexity were proposed in \cite{bell2020secure} and \cite{choi2020communication} requiring 3 rounds of interaction between the server and clients. In \cite{bonawitz2017practical}, the star topology of the communication network was replaced with a random subset of clients and secret sharing was only used for a subset of clients instead of all client pairs. Shamir's t-out-of-n Secret Sharing technique prevents the splitted subgroups from divulging any information about the original. In \cite{choi2020communication}, the proposed secure aggregation (CCESA) algorithm provided data privacy using substantially reduced communication and computational resources compared to other secured solutions. The key idea was to design the topology of secret-sharing nodes as a sparse random graph instead of the complete graph \cite{bonawitz2017practical}. The required resource of CCESA is reduced by a factor of at least $O(\sqrt{n / \log{n}})$ compared to \cite{bonawitz2017practical}.

\begin{table}[htb]
\footnotesize
\centering
\caption{Features and Benefits of Tree Topology}
\label{table:treebenfits}
\renewcommand{\arraystretch}{1.5}
\begin{tabularx}{.99\textwidth}{|p{.3\textwidth}|X|}
\hline
\textbf{Features} & \textbf{Benefits} \\ \hline
Clustered clients & Adaptive strategies of in-cluster communications based on cluster's condition.  \\
\hline
Configurable cluster & Better scalability compared to star topology  \\
\hline
Configurable number of layers & Varying policies for client-edge and edge-cloud, and inter-layer aggregations. \\
\hline
\end{tabularx}
\end{table}

\begin{figure}[htb]
    \centering
    \includegraphics[width=0.36\linewidth]{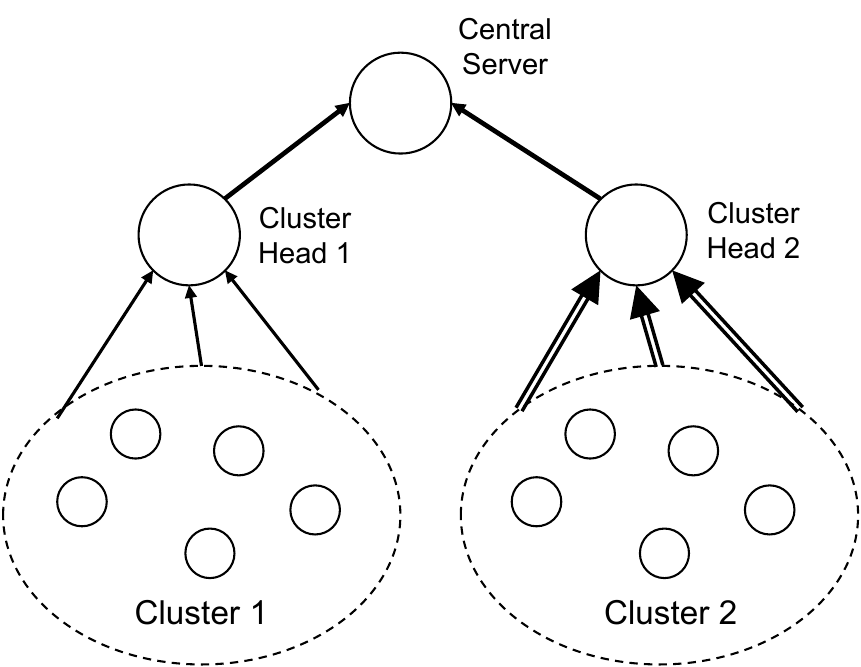}
    \includegraphics[width=0.36\linewidth]{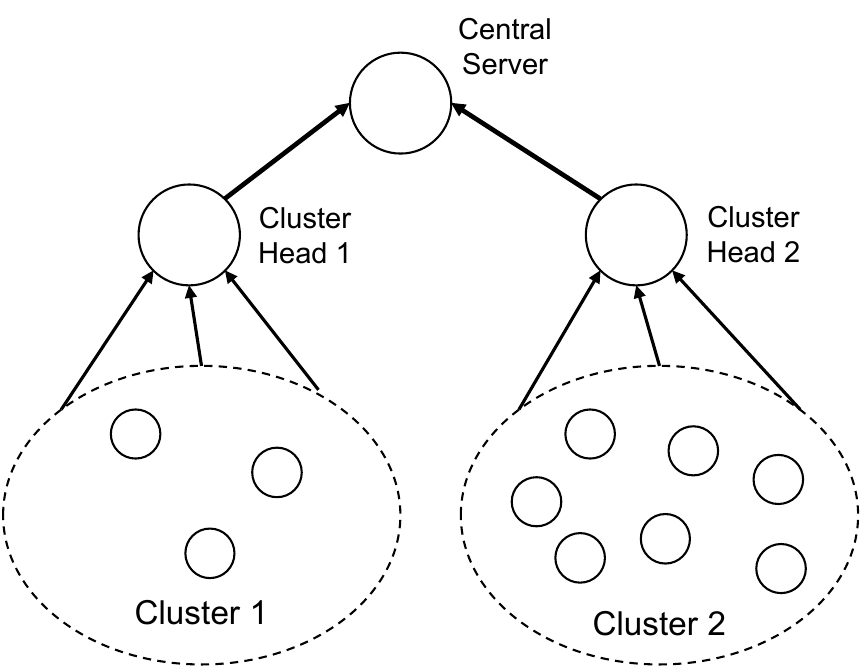}
    \caption{\textbf{Right: }FL with Tree topology enables varying communication costs in different clusters depending on their energy profiles. \textbf{Left: }The cluster size can be different for tree topology based on data distribution and other parameters of each cluster.}
    \Description{Right: A Central server connected with cluster head 1 and cluster head 2. Each cluster head has the same amount of devices in the cluster. Left: A Central server connected with cluster head 1 and cluster head 2. Each cluster head has a different amount of devices in the cluster.}
    \label{fig:hfl-benefit1}
\end{figure}


\begin{figure}[htb]
    \centering
    \includegraphics[width=0.6\linewidth]{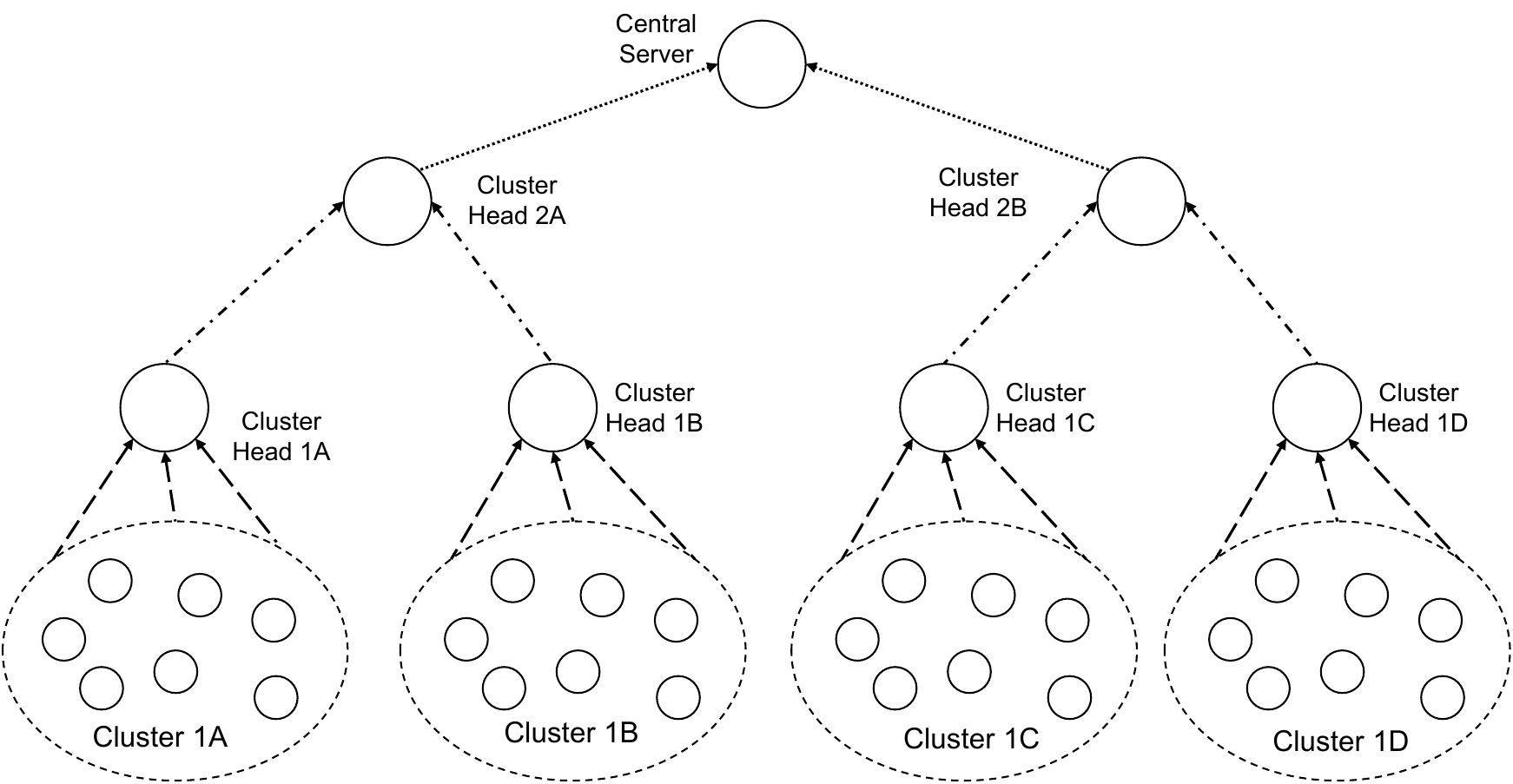}
    \caption{There can be arbitrary layers of clients in Tree topology FL system.}
    \Description{A Central server connected with cluster heads 2A and 2B. Cluster head 2A is connected with 1A and 1B. Cluster head 2B is connected with 1C and 1D. Each cluster has the same amount of devices.}
    \label{fig:hfl-benefit3}
\end{figure}

\subsection{Tree Topology}
There can be additional layers between the central server and edge devices. For instance, edge servers that connect edge devices and the central server can formulate one or multiple layers, making a tree-like topology with the highest level of the tree being the central server and the lowest level being edge devices.
Tree topologies must contain at least three levels. Otherwise, they are considered star topologies. Compared to traditional FL, tree topology helps overcome performance bottlenecks and single points of failure. We list the features and benefits of tree topology in Table \ref{table:treebenfits}.
In this section, we discuss applications and motivations for adopting tree topology. 
A review of state-of-the-art optimization frameworks and algorithms is presented. We show some visualization of tree topology structures and their benefits for FL in Fig. \ref{fig:hfl-benefit1}, Fig. \ref{fig:hfl-benefit1}, and Fig. \ref{fig:hfl-benefit3}. In the end, we cover some grouping strategies and privacy enhancement schemes. 

\begin{figure}[htb]
    \centering
    \includegraphics[width=0.55\linewidth]{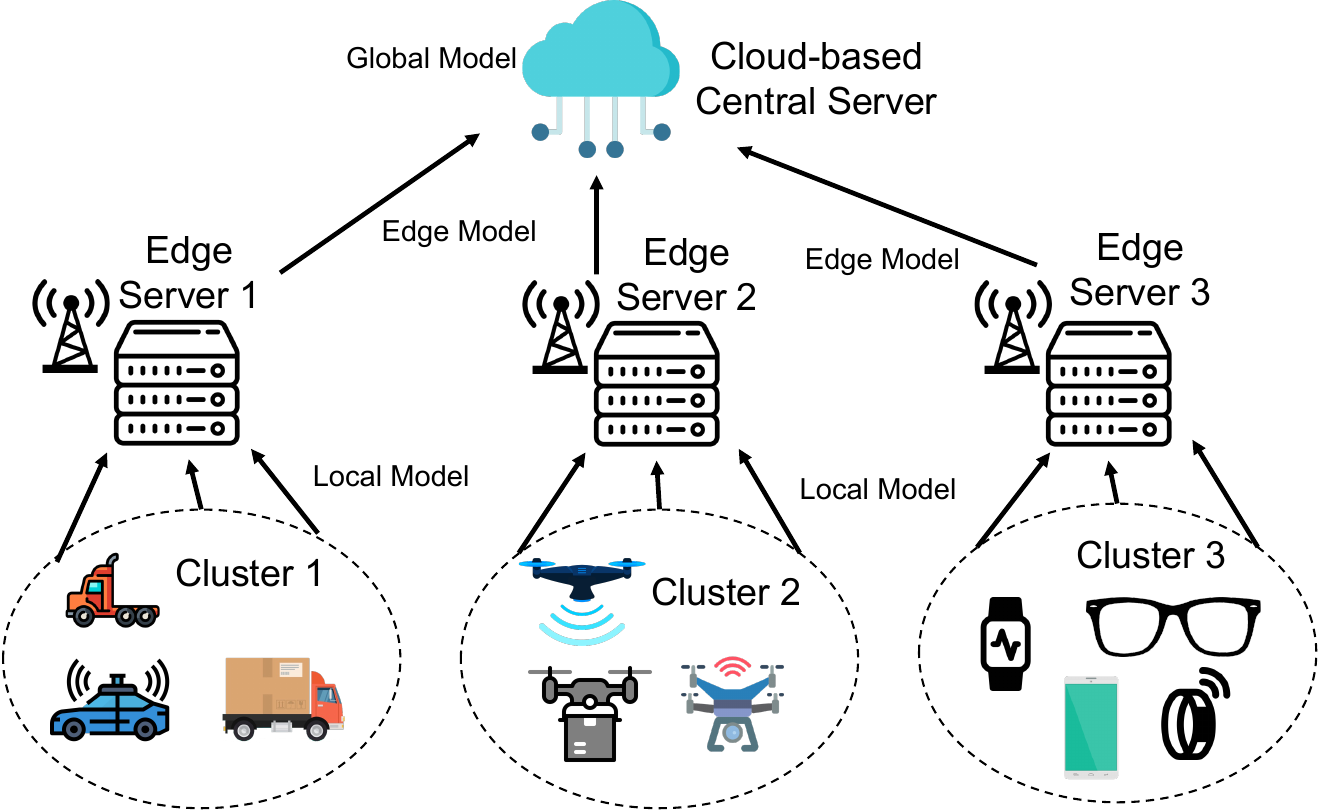}
    \caption{Hierarchical FL following tree topologies. Typically, the FL network has a tree structure with at least three tiers: the cloud tier, the edge tier, and the device tier.}
    \Description[Hierarchical model of FL with cloud and edge layers]{A Cloud-based Central Server houses the Global Model. Three edge servers are shown below the central server. Each edge server has a local cluster that includes devices like forklifts, trucks, drones, surveillance cameras, and wearable devices.}
    \label{fig:hfel}
\end{figure}

There are two major categories of FL studies in tree topology: hierarchical and dynamic. Hierarchical represents the classic two-tier hierarchy in topology design, while dynamic continues to follow the overall structure of the tree topology with some modifications. In the following sections, we organize the works that use the tree topology based on their topics. We show a classic example of Hierarchical FL in Fig. \ref{fig:hfel}. We list the features and benefits of tree topology in Table \ref{table:treebenfits}. We highlight some of the works using tree topology in Table \ref{table:hierarchicalHW} and Table \ref{table:DynamicHW}.

\subsubsection{Typical tree topology FL}  
\citet{zhou2021two} proposed a typical use of end-edge-cloud federated learning framework in 6G. The authors integrated a convolutional neural network-based approach specially perform hierarchical and heterogeneous model selection and aggregation using individual vehicles and RSUs at the edge and cloud level. Evaluation results showed an overall better outperform in learning accuracy, precision, recall, and F1 score compared to other state-of-the-art methods in 6G network settings.

\citet{yuan2020hierarchical} designed a LAN-based hierarchical federated learning platform to solve the communication bottleneck. The authors regarded that existing FL protocols have a critical communication bottleneck in a federated network coupled with privacy concerns, usually powered by a wide-area network (WAN). Such a WAN-driven FL design led to significantly higher costs and much slower model convergence. An efficient FL protocol was proposed to create groups of LAN domains in P2P mode without an intermediate edge server which involved a hierarchical aggregation mechanism using Local Area Network (LAN), as it had abundant bandwidth and almost negligible cost compared to WAN.

The benefits of training data aggregation at the edge in HFL were acknowledged by \citet{deng2021share}. When comparing HFL and cloud-based FL, $\kappa_e$ and $\kappa_c$ were defined as the aggregation frequency at the edge and the cloud, respectively. It was concluded from their research that in the HFL framework, with fixed $\kappa_e$ and $\kappa_c$,\textit{uniform distribution} of the training data at the edge significantly enhanced FL performance and reduced the rounds of communications. The original problem was first divided into two sub-problems to minimize the per-round communication cost and mean Kullback–Leibler divergence (KLD) of edge aggregator data. Then two lightweight algorithms were developed, adopting a heuristic method formulating a topology encouraging a uniform distribution of training data.

Briggs et al. pointed out in \cite{briggs2020federated} that in reality, most data was distributed in a non-IID fashion. These fashions included feature distribution skew, label distribution skew, and concept shifts. In such cases, most FL methods suffered accuracy loss. They introduced a hierarchical clustering step (FL+HC) to separate clusters of clients by the similarity of their local updates to the global joint model. Then multiple models targeted toward groups of clients were preferred. The empirical study showed that FL+HC allowed the training to converge in fewer communication rounds with higher accuracy.

\subsubsection{Optimization: Trade-off among energy cost, communication delay, model accuracy, data privacy}  
\citet{liu2020client} proposed a client-edge-cloud hierarchical learning system that reduced communication with the cloud by trading off between the client-edge and edge-end communication costs. This is achieved by leveraging the edge server’s ability to exchange local updates with clients constantly. There are two types of data collection rounds: one is from the client to the edge servers, and the other is from the edge servers to the cloud. The proposed FL algorithm Hierarchical Federated Averaging (HierFAVG) extends from the classic FAVG algorithm. Under the HierFAVG architecture, after the local clients finish $k_1$ rounds of training, each corresponding edge server aggregates its client’s model for $k_2$ rounds of aggregation, and the cloud server then aggregates all the edge servers’ models. Compared to the traditional systems following the star topology, this tree topology-based architecture greatly reduced the total communication rounds with the cloud server. Standard MNIST and CIFAR-10 datasets were used for the experiment. Additional two non-IID cases for MNIST were also considered. Experiments showed promising results on reduced communication frequency and energy consumption. When the overall communication ($k_1 k_2$) is fixed, fewer rounds of local updates ($k_1$) and more communication rounds with the edge will result in faster training which effectively reduces the number of computation tasks on the local clients. For the case of IID data on the edges, fewer communication rounds with the cloud server will not result in a decrease in performance as well. On energy consumption, with moderately increased communication with clients and edge, the energy consumption decreases. However, excessive communication between edge servers and clients will result in extra energy consumption. Therefore, a balance of overall communication ($k_1 k_2$) is needed to minimize energy consumption.

\begin{table}[htb]
\footnotesize
\centering
\caption{Highlighted works - Tree Topology - Hierarchical}
\label{table:hierarchicalHW}
\renewcommand{\arraystretch}{1.5}
\begin{tabularx}{.99\textwidth}{|p{1.5cm}|p{3.2cm}|X|X|}
\hline
\textbf{FL Type} & \textbf{Baselines \& Benchmarks} & \textbf{Key Findings } & \textbf{Performance}\\ \hline

\multirow{20}{45pt}{Hierarchical}
 & Hierarchical FL using CNN and mini-batch SGD with MNIST and CIFAR-10 under non-IID setting & Vanilla hierarchical FL, ignores heterogeneous distribution & Reduced communication, training time, and energy cost with the cloud. Also achieved efficient client-edge communication. \cite{liu2020client} \\  \cline{2-4}
 & Resource allocation methods and Fedavg with MNIST and FEMNIST & Multiple edge servers can be accessed by the device. Optimize device computation capacity and edge bandwidth allocation  & Better global cost-saving, training performance, test\&training accuracy, and lower training loss than FedAvg \cite{luo2020hfel} \\ \cline{2-4} 
 & Binary tree \& static saturated structure and FSVRG \& SGD algorithm with MNIST & Using the layer-by-layer approach, more edge nodes can be included in the model aggregation
& Scalability (time cost increases logarithmically rather than linearly in traditional FL), reduced bandwidth usage and time-consuming \cite{cao2021layered} \\  \cline{2-4}
 & Uniform, gradient-aware, and energy-aware scheduling with MNIST & Optimize scheduling and resource allocation by striking a balance between 3 scheduling schemes & Outperformed the baselines if $\lambda$ is chosen properly. Otherwise slightly better or worse performance \cite{wen2022joint} \\  \cline{2-4}
 & FedAvg plus SGD using CNN with MNIST & Both the central server and the edge servers are responsible for global aggregation & Reduced global communication cost, model training time and energy consumption  \cite{ye2020edgefed} \\  \cline{2-4}
 & RF, CNN, and RegionNet with BelgiumTSC & Classic hierarchical FL in 5G\&6G settings for object detection & Faster convergence and better learning accuracy for 6G supported IoV applications \cite{zhou2021two}\\ \cline{2-4}
 & FedAvg with imbalanced EMNIST and CINIC-10, CIFAR-10 & Relieved global and local imbalance of training data; recover accuracy & Significantly reduced communication cost and achieved better accuracy on imbalanced data \cite{duan2020self} \\ \cline{2-4}
 & FedAvg with MNIST and FEMNIST under IID \& non-IID settings & A clustering step was introduced to determine client similarity and form subsets of similar clients & Fewer communication rounds especially for some non-IID settings. Allowed more clients to reach target accuracy \cite{briggs2020federated} \\ \cline{1-4}
 
\hline
\end{tabularx}
\end{table}

\citet{luo2020hfel} also introduced a Hierarchical Federated Edge Learning (HFEL) framework to jointly minimize energy consumption and delay.
The authors formulated a joint computation and communication resource allocation problem for global cost optimization, which considered minimizing system-wide energy and delay within one global iteration, denoted by the following equation:
\begin{equation}
    E = \sum_{i \in \kappa } \left( E^{cloud}_{i} + E^{edge}_{S_i} \right ),
\end{equation}

\begin{equation}
    T = \max_{i \in \kappa } \left\{ T^{cloud}_{i} + T^{edge}_{S_i} \right\},
\end{equation}
where $E$ represented the total energy consumed by each edge server $i \in \kappa$ aggregating data and each sets of devices $S_i$ for edge server $i$ uploading models.
$T$ was defined as the total delay with those introduced by the edge servers to the cloud, denoted by $T^{cloud}_{i}$, and by the sets of devices uploading models, denoted by $T^{edge}_{S_i}$.
The optimization was to jointly minimize $E$ and $T$ with varying weights.
A resource scheduling algorithm was developed based on the model, which relieved the core network transmission overhead and enabled great potential in low-latency and energy-efficient FL.

Cao et al. proposed a federated learning system \cite{cao2021layered} with an aggregation method using the topology of the edge nodes to progress model aggregation layer by layer, specifically allowing child nodes on the lower levels to complete training first and then upload results to the higher node. Compared to the traditional FL architecture, where all the end devices connect to the same server, the proposed layered and step-wise approach ensures that only one gradient data is transmitted in a link at most. The simulation result shows better scalability where the time cost increases logarithmically rather than linearly in traditional FL systems.

Another joint optimization strategy that investigated the trade-off between computation cost and accuracy was presented by \citet{wen2022joint} for hierarchical federated edge learning (H-FEEL), where an optimization approach was developed to minimize the weighted sum of energy consumption and gradient divergence. 
The innovative contributions included three phases: local gradient computing, weighted gradient uploading, and model updating.

\citet{ye2020edgefed} proposed EdgeFed featuring the trade-off between privacy and computation efficiency.
In the EdgeFed scheme, split training was applied to merge local training data into batches before being transmitted to the edge servers.
Local updates from mobile devices were partially offloaded to edge servers where more computational tasks are assigned, reducing the computational overhead to mobile devices, which can focus on the training of \textit{low layers}. 
In the EdgeFed algorithms, each iteration included multiple split training between $K$ edge devices and corresponding edge servers and a global aggregation between $m$ edge servers and the central server. %
The edge device $k$ performed calculations with local data on low layers of the multi-layer neural network model.
After receiving the outputs of low layers from all edge devices, the edge server $m$ aggregated all data received into a larger matrix $x^{m}_{pool}$, which was then taken as the input of the remaining layers:

\begin{equation}
    x_{pool}^{m} \leftarrow \left [ x_{conv}^{1}, x_{conv}^{2}, \cdots x_{conv}^{k}, \cdots, x_{conv}^{K} \right ]
\end{equation}

As the updates to the central server were narrowed down between edge servers and the central server, with more computational power on the edge server compared to edge devices, the overall communication costs were reduced. However, the transferred data processed by the low layer of the model may be a threat to privacy because the edge server may be able to restore the original data.

\subsubsection{Dynamic Topology}  
\citet{mhaisen2021optimal} proposed that the tree topology can be dynamic with a dense edge network. Edge devices may pair with different edge servers in different rounds of data aggregation. When participants change either due to the client selection strategy \cite{reisizadeh2020fedpaq, wang2021device} or participants entering or exiting the network \cite{kourtellis2020flaas}, the topology will subsequently change. The authors argued that user equipment (UE) had access to more than one edge server in dense networks and increased the mobility of UE. Choosing the best edge server resulted in their proposed UE-edge assignment solutions. The user assignment problem was formalized in HFL based on the analysis of learning parameters with non-IID data.

\citet{kourtellis2020flaas} explored the possibility of collaborative modeling across different 3rd-party applications and presented federated learning as a service (FLaaS), a system allowing 3rd-party applications to create models collaboratively. A proof-of-concept implementation was developed on a mobile phone setting, demonstrating 100 devices working collaboratively for image object detection.
\begin{table}[htb]
\footnotesize
\centering
\caption{Highlighted works - Tree Topology - Dynamic}
\label{table:DynamicHW}
\renewcommand{\arraystretch}{1.5}
\begin{tabularx}{.99\textwidth}{|p{1.2cm}|p{3.2cm}|X|X|}
\hline
\textbf{FL Type} & \textbf{Baselines \& Benchmarks} & \textbf{Key Findings } & \textbf{Performance}\\ \hline

\multirow{16}{45pt}{Dynamic}
 & FedAvg using Random and heuristic sampling with MNIST and F-MNIST & Able to offload data from non-selected devices to selected devices during training & Significant improvements in datapoints processed, training speed, and model accuracy \cite{wang2021device} \\ \cline{2-4}
 & FedAvg using F-Fix and F-Opt with CNN on MNIST &Flexible system topology that optimizes computing speed and transmission power & Accelerated the federated learning process, and achieved a higher energy efficiency \cite{huang2021joint} \\ \cline{2-4}
 & WAN-FL using CNN with FEMNIST and CelebA under non-IID settings & Dynamic device selection based on the network capacity of LAN domains. Relied heavily on manual parameter tuning & Accelerate training process, saved WAN traffic, and reduced monetary cost while preserving model accuracy \cite{yuan2020hierarchical} \\ \cline{2-4}
 & FedAvg, TiFL, FedAsync with FMNIST, CIFAR-10, Sentiment140 & Models were updated synchronously with clients of the same tier and asynchronously with the global model across tiers 
 & Faster convergence towards the optimal solution, improved prediction performance, and reduced communication cost \cite{chai2020fedat} \\\cline{2-4}
 & Cloud-based FL (C-FL), Cost only CPLEX (CC), Data only greedy (DG) with MNIST and CIFAR-10 & As opposed to an edge server, groups of distributed nodes are used for edge aggregation & Improved FL performance at a very low communication cost, provided a good balance between learning performance and communication costs \cite{deng2021share} \\ \cline{2-4}
 & Traditional FL(TFL) low \& high power mode with MNIST under IID \& non-IID settings &Based on the status of their local resources, clients are assigned to different subnetworks of the global model & Outperformed TFL in both low \& high power modes, especially in low power. Reliable in dynamic wireless communication environments \cite{yu2021toward} \\ \cline{1-4}

 \hline
\end{tabularx}
\end{table}
FedPAQ was proposed in \cite{reisizadeh2020fedpaq} as a communication-efficient federated learning method with periodic averaging and quantization. FedPAQ’s first key feature was to run local training before synchronizing with the parameter server. The second feature of FedPAQ was to capture the constraint on the availability of active edge nodes by allowing partial node participation, leading to better scalability and a smaller communication load. The third feature of FedPAQ was that only a fraction of device participants sent a quantized version of their local information to the server during each round of communication, significantly reducing the communication overhead.

The device sampling in Heterogeneous FL was studied in \cite{wang2021device}. The authors noticed that there may be significant overlaps in the local data distribution of devices. Then a joint optimization was developed with device sampling aiming at selecting the best combination of sample nodes and data offloading configurations to maximize FL training accuracy with network and device capability constraints.

\citet{huang2021joint} proposed a novel topology-optimized federated edge learning (TOFEL) scheme where any devices in the system received and aggregated their own gradients and then passed them to other devices or edge servers for further aggregation. The system acted as a hierarchical FL topology with adjustable gradient uploading and aggregation topology. The authors formulated a joint topology and computing speed optimization as a mixed-integer nonlinear program (MINLP) problem aims at minimizing energy consumption and latency. A penalty-based successive convex approximation (SCA) method was developed to transform the MINLP into an equivalent continuous optimization problem which demonstrates that the proposed TOFEL scheme speed up the federated learning process while consuming less energy.

Duan et al. \cite{duan2020self} focused on the imbalanced data distribution in mobile systems, which led to model biases. The authors built a self-balancing FL framework called Astraea to alleviate the imbalances with a mediator to reschedule the training of clients. The methods included Z-score-based data augmentation and mediator-based multi-client rescheduling. The Astraea framework consisted of three parts: FL server, mediator, and clients. 

\subsubsection{Grouping strategy and privacy enhancement}  
\citet{he2022improving} proposed a grouping mechanism called Auto-Group, which automatically generated grouped users using an optimized Genetic Algorithm without the need to specify the number of groups. The Genetic Algorithm balanced the data distribution of each group to be as close as the global distribution.

FedAT was a method with asynchronous tiers under non-IID data proposed by \citet{chai2020fedat}. FedAT had the topology organized in tiers based on the response latencies of edge devices. For intra-tier training, the synchronous method was used as the latencies are similar. For cross-tier training, the asynchronous method was used. The clients were split into tiers based on their training speed with the help of the tier-based module, allowing faster clients to complete more local training while using server-side optimization to avoid bias. By bridging the synchronous and asynchronous training through tiering, FedAT minimized the straggler effect with improved convergence speed and test accuracy. FedAT used a straggler-aware, weighted aggregation heuristic to steer and balance the training for further accuracy improvement. FedAT compressed the uplink and downlink communications using an efficient, polyline-encoding-based compression algorithm, therefore minimizing the communication cost. Results showed that FedAT improved the prediction performance by up to 21.09\%, and reduced the communication cost by up to 8.5 times compared to the state-of-the-art FL methods.

With the two typical FL scenarios in MEC, i.e., virtual keyboard and end-to-end autonomous driving, Yu et al. proposed a neural-structure-aware resource management approach \cite{yu2021toward} for FL. The mobile clients were assigned to different subnetworks of the global model based on the status of local resources. 

\citet{wainakh2020enhancing} discussed the implications of the hierarchical architecture of edge computing for privacy protection. The topology and algorithm enabled by hierarchical FL (HFL) may help enhance privacy compared to the original FL. These enhancements included flexible placement of defense and verification methods within the hierarchy and the possibility of employing trust between users to mitigate several threats. The methods linked to HFL were illustrated, such as sampling users, training algorithms, model broadcasting, and model updates aggregation. Group-based user update verification could also be introduced with HFL. Flexible applications of defense methods were available in HFL because of the hierarchical nature of the network topology.

\subsection{Decentralized\textbackslash Mesh Topology}

Decentralized\textbackslash Mesh topology is a network topology where all end devices are inter-connected to each other in a local network \cite{lalitha2018fully, shi2021over, xing2020decentralized, liu2021decentralized, chen2021dacfl, zehtabi2022decentralized}. In recent studies, mesh topologies are commonly used in FL systems. Decentralized approaches like peer-to-peer (P2P) or device-to-device (D2D) FL fall under the Mesh Topology. Many existing FL systems still rely on a centralized/cloud server for model aggregation \cite{konevcny2016federated}. The decentralized approach is sometimes regarded as a poor alternative to the centralized method when a centralized server is not feasible. This section covers three major FL systems using fully decentralized approaches. In Table \ref{table:DecentralizedHW}, we highlighted some works that utilized the decentralized topology.


The performance of decentralized FL algorithms was discussed in \cite{lian2017can}. Theoretical analysis on a D-PDGD algorithm was conducted to prove the possibility of a decentralized FL algorithm outperforming centralized FL algorithms. With comparable computational complexity, the decentralized FL algorithm required much less communication cost on the busiest node. D-PSGD could be one order of magnitude faster than its well-optimized centralized counterparts.

\citet{lu2020federated} developed a vehicular FL scheme based on a sub-gossip update mechanism along with a secure architecture for vehicular cyber-physical systems (VCPS). The P2P vehicular FL scheme used random sub-gossip updating without a curator, which enhanced security and efficiency. The aggregation process was done in each vehicle asynchronously. The data retrieval information was registered on nearby RSUs as a distributed hash table (DHT). All related vehicles were searched for the DHT before FL started.

Gossip learning \cite{hegedHus2016robust, ormandi2013gossip} was compared with FL in \cite{hegedHus2019gossip} as an alternative, where training data also remained at the edge devices, but there was no central server for aggregation. Gossip learning can be seen as a variation of the mesh topology. Nodes exchanged and aggregated models directly. No centralized servers meant no single-point failure and led to better scalability and robustness. The performance of gossip learning was generally comparable with FL and even better in some scenarios. The experiment was conducted using PEERSIM \cite{p2p09-peersim}
%

\begin{table}[htb]
\footnotesize
\centering
\caption{Highlighted works - Decentralized Topology}
\label{table:DecentralizedHW}
\renewcommand{\arraystretch}{1.5}
\begin{tabularx}{.99\textwidth}{|l|p{4cm}|X|}
\hline
\textbf{FL Type} & \textbf{Baselines \& Benchmarks} & \textbf{Key Findings } \\ \hline

\multirow{8}{45pt}{Decentralized Mesh}
 & Using 20 Newsgroups dataset integrating GBDT & Obtained high utility and accuracy, effective data leakage detection, near-real-time performance data leakage defending \cite{lu2020federated} \\  \cline{2-3}
 & FedAvg and FedGMTL using AGE and GAT wtih MoleculeNet & Train GNNs in serverless scenarios, outperformed Star FL even if clients can only communicate with few neighbors \cite{he2021spreadgnn} \\  \cline{2-3}
 & PENS, Random, Local, FixTopology, Oracle, IFCA, FedAvg with MNIST, FMNIST, and CIFAR10 & CNI was effective to match neighbors with similar objectives; directional communications helped to converge faster; robust in non-IID settings \cite{li2022mining}  \\  \cline{2-3}
 & FedAvg using ResNet-20 model with CIFAR-10 under IID \& non-IID settings & Provided an unbiased estimate of the model update to PS through relaying; optimized consensus weights of clients to improve convergence; compatible in different topologies \cite{yemini2022robust} \\ \cline{1-3}
 
\multirow{17}{45pt}{Decentralized Wireless}
 &  FedAvg, CDSGD, D-PSGD using CNN with MNIST, FMNIST, CIFAR-10 under IID \& non-IID settings & Outperformed in accuracy, less sensitive to the topology sparsity; similar performance for each user; viable on IID \& non-IID data under time-invariant topology \cite{chen2021dacfl} \\  \cline{2-3}
 & DSGD, TDMA-based, local SGD(no communication) with FMNIST & Over-the-air computing can only outperform conventional star topology implementations of DSGD \cite{xing2020decentralized} \\  \cline{2-3}
 & DOL and COL with SUSY and Room-Occupan & Worked better than DOL in row stochastic confusion matrix, usually outperformed COL in running time \cite{he2019central} \\  \cline{2-3}
 & FedAvg and gossip approach without segmentation with CIFAR-10 & Required the least training time to achieve given accuracy, more scalable, synchronization time significantly reduced \cite{hu2019decentralized} \\  \cline{2-3}
 & Gossip and Combo with FEMNIST and Synthetic data & Maximize bandwidth utilization by segmented gossip aggregation over the network; speed up training; maintain convergence\cite{jiang2020bacombo} \\  \cline{2-3}
 & DFL and C-SGD with MNIST, CIFAR-10 & Showed linear convergence behavior for convex objective, strong convergence guarantees for both DFL and C-DFL \cite{liu2021decentralized} \\  \cline{2-3}
 & FLS with MALC dataset using QuickNAT architecture & Enabled more robust training; similar performance with centralized approaches; generic method and transferable \cite{roy2019braintorrent} \\  \cline{2-3}
 & FedAvg with MNIST using CNN and LSTM & Improved convergence performance of FL especially when the model was complex and network traffic was high \cite{pinyoanuntapong2020fedair} \\  \cline{2-3}
 & Gossip and GossipPGA using LEAF with FEMNIST and Synthetic data & Reduced training time and maintained good convergence, whereas partial exchange significantly reduced latency \cite{jiang2020decentralised} \\  \cline{1-3}
 
\hline
\end{tabularx}
\end{table}

SpreadGNN was proposed in \cite{he2021spreadgnn} as a novel multi-task federated training framework able to operate with partial labels of client data for graph neural networks in a fully decentralized manner. The serverless multitask learning optimization problem was formulated, and Decentralized Periodic Averaging SGD (DPA-SGD) was introduced to solve the problem. The result shows that it is viable to train graph neural networks federated learning in a fully decentralized setting.

\citet{li2022mining} leveraged P2P communications between FL clients without a central server and proposed an algorithm formulating a decentralized, effective communication topology in a decentralized manner without assuming the number of clusters. To design the algorithm, two novel metrics were created for measuring client similarity. Another two-stage algorithm directed the clients to match same-cluster neighbors and to discover more neighbors with similar objectives. Theoretical analysis was included showing the effectiveness of the work compared to other P2P FL methods.

A semi-decentralized topology was introduced by \citet{yemini2022robust}, where a client was able to relay the update from its neighboring clients. A weighted update with both the client's own data and its neighboring clients' data was transmitted to the parameter server (PS). The goal was to optimize averaging weights to reduce the variance of the global update at the PS, as well as minimize the bias in the global model, eventually reducing the convergence time.

\subsubsection{Decentralized Topology in Wireless Networks}  
The mesh or decentralized FL topology has been explored in wireless networks \cite{shi2021over, yang2020federated,xing2020decentralized,gholami2022trusted}, as the wireless coverage of P2P or D2D devices overlaps with each other, plus no centralized server has been provided.

Trust was treated as a metric of FL in \cite{gholami2022trusted}. The trust was quantified upon the relationship among network entities according to their communication history. Positive contributions to the model were interpreted as an increment of trust, and vice versa. 

Shi et al. proposed over-the-air FL \cite{shi2021over} over wireless networks, where over-the-air computation (AirComp) \cite{yang2020federated} was adopted to facilitate the local model consensus in a D2D communication manner.

\citet{chen2021dacfl} considered the deficiency of high divergence and model average necessity in previous decentralized FL implementations like CDSGD and D-PSGD. They devised a decentralized FL implementation called DACFL \cite{chen2021dacfl} which adapts more to non-ideal network topology. DACFL allows individual users to train their own model with their own training data while exchanging the intermediate models with neighbors using FODAC (first-order dynamic average consensus) to negate potential over-fitting problems discretely without a central server during training.

Xing et al. considered a network of wireless devices sharing a common fading wireless channel for deploying  FL \cite{xing2020decentralized}. Each device held a generally distinct training set, and communication typically took place in a D2D manner. In the ideal case, where all devices within their communication range could communicate simultaneously and noiselessly, a standard protocol guaranteed the convergence to an optimal solution of the global empirical risk minimization problem under convexity and connectivity assumptions was called the Decentralized Stochastic Gradient Descent (DSGD). DSGD integrated local SGD steps with \textit{periodic consensus averages} that required communication between neighboring devices. Wireless protocols were proposed for implementing DSGD by accounting for the presence of path loss, fading, blockages, and mutual interference.

He et al. explored the use cases of FL in social networks where centralized FL was not applicable \cite{he2019central}. Online Push-Sum (OPS) method was proposed to leverage trusted users for aggregations. OPS offered an effective tool to cooperatively train machine learning models in applications where the willingness to share is single-sided.

Lalitha et al. considered the problem of training models in fully decentralized networks. They proposed a distributed learning algorithm \cite{lalitha2018fully} where users aggregated information from their on-hop neighbors to learn a model that best fits their observations to the entire network with small probabilities of error. 

Savazzi et al. proposed a fully distributed, serverless FL approach to massively dense and fully decentralized networks \cite{savazzi2020federated}. Devices are independently trained based on local datasets received from neighbors. Then devices forwarded the model updates to their one-hop neighbors for a new consensus step, extending the method of gossip learning. Both model updates and gradients were iteratively exchanged to improve convergence and minimize the rounds of communications.

Combo, a decentralized federated learning system based segmented gossip approach was presented in \cite{hu2019decentralized} to split the FL model into segmentations. A worker updated its local segmentation with k other workers, where k was much smaller than the total number of workers. Each worker stochastically selected a few other workers for each training iteration to transfer model segmentation. Replication of models was also introduced to ensure that workers had enough segmentation for training purposes.

Jiang et al. proposed Bandwidth Aware Combo (BACombo) \cite{jiang2020bacombo} with a segmented gossip aggregation mechanism that makes full use of node-to-node bandwidth to speed up the communication time. Besides, a bandwidth-aware worker selection model further reduces the transmission delay by greedily choosing the bandwidth-sufficient worker. The convergence guarantees were provided for BACombo. The experimental results on various datasets demonstrated that the training time was reduced by up to 18 times that of baselines without accuracy degradation.

The work in \cite{liu2021decentralized} focused on balancing between communication-efficiency and convergence performance of Decentralized federated learning (DFL). The proposed framework performed both multiple local updates and multiple inter-node communications periodically, unifying traditional decentralized SGD methods. Strong convergence guarantees were presented for the proposed DFL algorithm without the assumption of a convex objective function. The balance of communication and computation rounds was essential to optimize decentralized federated learning under constrained communication and computation resources. To further improve the communication efficiency of FL, compressed communication was applied to DFL, which exhibited linear convergence for strongly convex objectives.

BrainTorrent \cite{roy2019braintorrent}  was proposed to perform FL for medical centers without the use of a central server to protect patient privacy. As the central server required trust from all clients, which was not feasible for multiple medical organizations, BrainTorrent presented a dynamic peer-to-peer environment. All medical centers directly interact with each other, acting like a P2P network topology. Each client maintains its model version of the model and the last versions of models it used during merging. By sending a ping request, a client receives responses from other clients with their latest model versions and subsets of the models. The client then merged the models received by weighted averaging to generate a model.

FedAir \cite{pinyoanuntapong2020fedair} explored enabling FL over wireless multiple-hop networks, including the widely deployed wireless community mesh networks. Wireless multi-hop FL system consists of a central server as an aggregator with multi-hop link wireless to edge servers as workers. According to the authors, multi-hop FL faced several challenges, including slow convergence rate, prolonged per-round training time and potential divergence of synchronous FL, and difficulties in model-based optimization for multiple hops.

Jiang et al. proposed gradient partial level decentralized federated learning (FedPGA) \cite{jiang2020decentralised} aiming to improve on traditional star topology FL’s high training latency problem in real-world scenarios. The authors used a partial gradient exchange mechanism to maximize the bandwidth to improve communication time, and an adaptive model updating method to adaptively increase the step size. The experimental results showed up to 14 × faster training time compared to baselines without compromising accuracy. 

\subsubsection{Routing in Decentralized Topology}  
A topology design problem for cross-silo FL was analyzed in \cite{marfoq2020throughput} due to traditional FL topology designs being inefficient in the cross-silo settings. They proposed algorithms that find the optimal topology using the theory of max-plus linear systems. By minimizing the duration of communication rounds or maximizing the largest throughput, they were able to find the most optimal topology design that significantly shortens training time.

Secco et al. proposed Blaster \cite{sacco2020federated}, a federated architecture for routing packets within a distributed edge network, to improve the application’s performance and allow scalability of data-intensive applications. A path selection model was proposed using Long Short Term Memory (LSTM) to predict the optimal route. Initial results were shown with a prototype deployed over the GENI testbed. This approach showed that communications between SDN controllers could be optimized to preserve bandwidth for the data traffic. 

In the Cross-device FL scenario, Ruan et al. studied flexible device participation in FL \cite{ruan2021towards}. The authors assumed that it was difficult to ensure that all devices were available during the entire training in practice. It could not guarantee that devices would complete their assigned training tasks in every training round as expected. Specifically, the research incorporated four situations: in-completeness where devices submitted only partially completed work in a round, inactivity where devices did not complete any updates or respond to the coordinator at all, early departures where existing devices quit the training without finishing all training rounds, and late arrivals where new devices joined after the training has already started.

In \cite{nguyen2021deep}, a Federated Autonomous Driving network (FADNet) was designed to improve FL model stability, ensure convergence, and handle imbalanced data distribution problems. The experiments were conducted with a dense topology called the Internet Topology Zoo (Gaia) \cite{knight2011internet}. 

A federated learning on a fully decentralized network problem was analyzed in \cite{kavalionak2021impact}, particularly on how the convergence of a decentralized FL system will be affected under different network settings. Several simulations were conducted with different topologies, datasets, and machine learning models. The end results suggested that scale-free and small-world networks are more suitable for decentralized FL and a hierarchical network has convergence speed with trade-offs.

\subsubsection{Blockchain-based Topology}  
\begin{figure}[htb]
    \centering
    \includegraphics[width=0.55\linewidth]{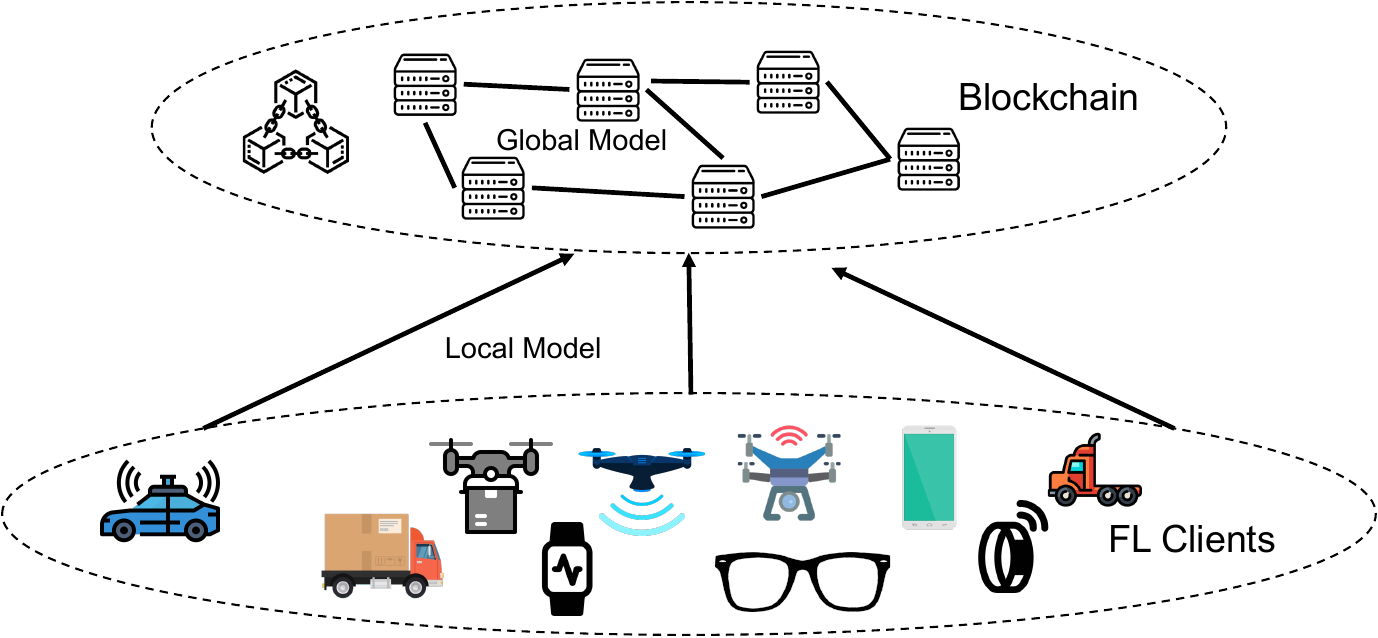}
    \caption{FL with blockchain as distributed ledgers to increase the availability of the central server.}
    \Description[FL with blockchain diagram]{The global Model is connected by blockchain blocks. Devices like forklifts, delivery trucks, drones, wearable glasses, and smartwatches can participate the FL process.}
    \label{fig:fl-blockchain}
\end{figure}

As one of the decentralized methods for FL, numerous blockchain-based topologies have been presented in previous works. The blockchains combined with FL aim at replacing the central server for generating the global model. Fig. \ref{fig:fl-blockchain} displays a typical blockchain-based FL topology. The highlighted works of blockchain-based topology are shown in Table \ref{table:BlockchainHW}. The potential benefits of introducing blockchains in FL systems include the following:

\begin{itemize}
    \item Placement of model training
    \item Incentive mechanism to attract more participants
    \item Decentralized privacy
    \item Defending poison attacks
    \item Cross verification
\end{itemize}

FLChain was proposed in \cite{majeed2019flchain} to enhance the reliability of FL in wireless networks for separate channel selection used by FL model uploading and downloading, where local model parameters were stored as blocks on a blockchain as an alternative to a central aggregation server and the edge devices provided network resources to the resource-constraint mobile devices and served as nodes in the blockchain network of FLchain. Similar to most blockchain-based FL frameworks, FLChain had the blockchain network above the edge devices for channel registration and global model updates.

A blockchain-based federated learning framework with committee consensus (BFLC) was proposed in \cite{li2020blockchain} to reduce the amount of consensus computing and malicious attacks. The alliance blockchain was used to manage FL nodes for permission control. Different from the traditional FL process, there is an additional committee between the training nodes and the central server for updates selection. In each round of FL, updates were validated and packaged by the selected committee, allowing the most honest nodes to improve the global model continuously. A small number of incorrect or malicious node updates will be ignored to avoid damaging the global model. Nodes can join or leave at any time without damaging the training process. The blockchain acted as a distributed storage system for persisting the updates.

\begin{table}[htb]
\footnotesize
\centering
\caption{Highlighted works - Blockchain Topology}
\label{table:BlockchainHW}
\renewcommand{\arraystretch}{1.5}
\begin{tabularx}{.99\textwidth}{|l|p{4cm}|X|}
\hline
\textbf{FL Type} & \textbf{Baselines \& Benchmarks} & \textbf{Key Findings } \\ \hline
\multirow{16}{45pt}{Blockchain}
 & basic FL and stand-alone training framework with FEMNIST & Higher resistance to malicious nodes, mitigate the influence of malicious central servers or nodes \cite{li2020blockchain} \\  \cline{2-3}
 & FL-Block with CIFAR-10, FASHION-MINIST & Fully capable to support big data scenarios, particularly fog computing applications, provides decentralized privacy protection while preventing a single point of failure \cite{qu2020decentralized}  \\  \cline{2-3}
 & Leaf with Ethereum as the underlying blockchain, tested logistic regression (LR) and NNs models & Incentive mechanisms encouraged clients to provide high-quality training data, communication overhead can be significantly reduced when the data set size is extremely large \cite{zhang2020blockchain} \\  \cline{2-3}
 & Integrate FL in consensus process of permissioned blockchain with Reuters and 20 newsgroups dataset & Increased efficiency of data sharing scheme by improving utilization of computing resources; secure data sharing with high utility and efficiency\cite{lu2019blockchain} \\  \cline{2-3}
 & Provided assistance to home appliance manufacturers using FL to predict future customer demands and behavior with MNIST & Created an incentive program to reward participants while preventing poisoning attacks from malicious customers; communication costs are small compared with wasted training time on mobile.  \cite{zhao2020privacy} \\  \cline{2-3}
 & Evaluation were based on 3GPP LTE Cat. M1 specification & Allowed autonomous vehicles to communicate efficiently, as it exploited consensus mechanisms in blockchain to enable oVML ML without centralized server \cite{pokhrel2020federated} \\  \cline{2-3}
 & FL with Multi-Krum \& DP under position attacks using Credit Card dataset and MNIST & Scalable; fault-tolerant; defend against known attacks; capable of protecting the privacy of client updates and maintaining the performance of the global model with 30\% adversaries \cite{shayan2020biscotti} \\  \cline{1-3}
\hline
\end{tabularx}
\end{table}

Qu et al. developed FL-Block \cite{qu2020decentralized} to allow the exchange of local learning updates from end devices via a blockchain-based global learning model verified by miners. The central authority was replaced with an efficient blockchain-based protocol. The blockchain miners verified and stored the local model updates. A linear regression problem was presented with the objective of minimizing a loss function $f(\omega)$. An algorithm designed for block-enabled FL enabled block generation by the winning miner after the local model was uploaded to the fog servers. The fog servers received updates of global models from the blockchain.

A blockchain anchoring protocol was designed \cite{zhang2020blockchain} for device failure detection. Specifically, a blockchain anchoring protocol was designed which built custom Merkle trees with each leaf node representing a record of data. root onto blockchains to verify Industrial IoT (IIoT) data integrity efficiently.

In a similar research scenario of processing IIoT data, a permissioned blockchain \cite{polge2021permissioned} was used in \cite{lu2019blockchain} for recording IIoT data retrieval and data sharing transactions. The Proof of Training Quality (PoQ) was proposed to replace the original Proof of Work (PoW) mechanism for lower cost reaching consensus. A differential privacy preserved model was first incorporated into FL. Regarding the PoQ, the committee leader was selected according to the trained model by prediction accuracy, measured by the mean absolute errors (MAEs):

\begin{equation}
    {MAE}(m_i) = \frac{1}{n} \sum_{i = 1}^{n} \left| y_i - f(x_i) \right|,
\end{equation}
where $f(x_i)$ denoted the prediction value of model $m_i$ and $y_i$ was the observed value. The consensus process started with the election of the committee leader with the lowest ${MAE}^u$ by voting. This leader was then assigned to drive the consensus process. The trained models were circulated among the neighboring committee nodes, denoted by $P_i$, of a committee node $P_j$, leading to the MAE for $P_j$ to be

\begin{equation}
    MAE^{u}(P_j) = \gamma \cdot {MAE}(m_j) + \frac{1}{n} \sum_{i = 1}^{n} {MAE}(m_i),
\end{equation}
where ${MAE}(m_j)$ was the locally trained model weighted by $\gamma$, and ${MAE}(m_i)$ referred to remotely trained models.

Zhao et al. \cite{zhao2020privacy} replaced aggregator nodes in traditional FL systems with blockchains for traceable activities. Customer data was selected to be sent to selected miners for averaging. One of the miners, selected as the leader, uploaded the aggregated model to the blockchain. More importantly, the authors proposed a normalization technique with differential privacy preservation.

Pokhrel et al. \cite{pokhrel2020federated} discussed blockchain-based FL (BFL) parameters for vehicular communication networking, considering local on-vehicle machine learning updates. The blockchain-related parameters were discussed via a mathematical framework, including the retransmission rate, block size, block arrival rate, block arrival rate, and frame sizes. The analytical results proved that tuning the block arrival rate was able to minimize the system delay.

Shayan et al. proposed a fully decentralized multi-party ML (Bitscotti) \cite{shayan2020biscotti} using blockchain and cryptographic emphasis on privacy-preserving. The training process of clients is stored in the blockchain ledger. Clients complete local training and the results are masked using a private noise. Then the masked updates go through a validation process as an extra layer of security. A new block is created for every new round of training. However, due to communication overhead, Bitscotti does not support large deep learning models. With a size of 200 peers, the test shows similar utility compared to traditional star topology federated learning.
\begin{table}[htb]
\footnotesize
\centering
\caption{Highlighted works - Minor Topologies}
\label{Table:MinorHW}
\renewcommand{\arraystretch}{1.5}
\begin{tabularx}{.99\textwidth}{|l|p{4cm}|X|}
\hline
\textbf{FL Type} & \textbf{Baselines \& Benchmarks} & \textbf{Key Findings } \\ \hline

\multirow{6}{45pt}{Ring}
 & FedAvg with MNIST, CIFAR-10 and CIFAR-100 & Improved bandwidth utilization, robustness, and system communication efficiency, reduce communication costs \cite{wang2021efficient} \\  \cline{2-3}
 & G-plain (Graph-based), R-plain (ring-based), and UBAR with MNIST and CIFAR-10 & Fast and computationally efficient; superior performance with SOTA in IID \& non-IID; achieve linear convergence rate; further scalability in parallel implementation  \cite{elkordy2022basil} \\  \cline{2-3}
 & LeNet and VGG11 with MNIST, FMNIST, EMNIST, CIFAR-10 & Achieved higher test accuracy in fewer communication rounds; faster convergence, robustness to non-IID dataset \cite{yang2021ringfed}  \\  \cline{1-3}
 
\multirow{2}{45pt}{Clique}
 & Fedavg with MNIST and CIFAR10 & Reduced gradient bias, convergence in heterogeneous data environments, reduction in edge and message numbers\cite{bellet2021d}  \\  \cline{1-3} 
 
\multirow{9.5}{45pt}{Fog}
 & FedAvg, HierFAVG, DPSGD with MNIST, FEMNIST, Synthetic dataset & Robust under dynamic topologies; fastest convergence rates under both static and dynamic topologies \cite{zhong2021p} \\  \cline{2-3}
 & FedAvg with MNIST, FEMNIST, Shakespeare & Gave smooth convergence curve; higher model accuracy; more scalable; communication-efficient  \cite{chou2021efficient} \\  \cline{2-3}
 & FL with full device participation and FL with one device sampled from each cluster with MNIST, F-MNIST & Better model accuracy; energy consumption; robustness against outages; favorable performance with non-convex loss functions \cite{lin2021semi}  \\  \cline{2-3}
 & Only Cloud, INC Solution, Non-INC, and INC LB & Reached near-optimal network latency; outperformed baselines; helped cloud node significantly decrease its network’s aggregation latency, traffic, and computing load \cite{dinh2021enabling} \\  \cline{2-3}
 & FedAsync and FedAvg with MNIST and CIFAR-10 & Reduced consumption of network traffic; faster converges; effective with non-IID data; dealing with staleness \cite{wang2022accelerating} \\  \cline{1-3}
 
\multirow{4}{45pt}{Semi-ring}
 & FedAvg with MNIST under non-IID setting & Improve communication efficiency, flexible and adaptive convergence \cite{tao2021efficient}  \\  \cline{2-3}
 & Astraea, FedAvg, HierFavg, IFCA, MM-PSGD, SemiCylic with FedShakespeare, MNIST & near-linear scalability; improved model accuracy \cite{lee2020tornadoaggregate} \\  \cline{1-3}

\hline
\end{tabularx}
\end{table}
\subsection{Minor Topologies}

In addition to the above-mentioned topologies, some minor topologies combine existing topologies or utilize niche new topologies that are not widely used. Although there are only a few studies on some minor topologies, these works can still provide valuable insight into the subject. The highlighted works of minor topology are shown in Table \ref{Table:MinorHW}.

\subsubsection{Ring Topology}
A ring-topology decentralized federated learning (RDFL) framework was proposed in \cite{wang2021efficient} for communication-efficient learning across multiple data sources in a decentralized environment. RDFL was inspired by the idea of ring-allreduce \footnote{https://andrew.gibiansky.com/blog/machine-learning/baidu-allreduce/} and applied a consistent hashing technique to construct a ring topology of decentralized nodes. An IPFS-based data-sharing scheme was designed as well to reduce communication costs.

RingFed \cite{yang2021ringfed} took advantage of the ring topology setup allowing clients to communicate with each other while performing preaggregation on clients to further reduce communication rounds. RingFed does not rely on the central server to perform model training tasks but uses the central server to assist in the passing of model parameters. The client only communicated with the central server when the set number of periods. In comparison to other algorithms, an additional step of recalculating all client parameters is added. Experimental results show that RingFed outperforms FedAvg in most cases and the results of training are optimized on non-IID data as well.

\citet{elkordy2022basil} proposed Basil, a fast and computationally efficient Byzantine robust algorithm for decentralized (serverless) training systems. In particular, the key aspect of their work is that it considers the decentralized FL and leverages the logical ring topology among nodes. Basil has also proven to achieve a linear convergence rate and further scalability in parallel implementation.

\subsubsection{Clique Topology}
\begin{figure}[htb]
    \centering
    \includegraphics[width=0.55\linewidth]{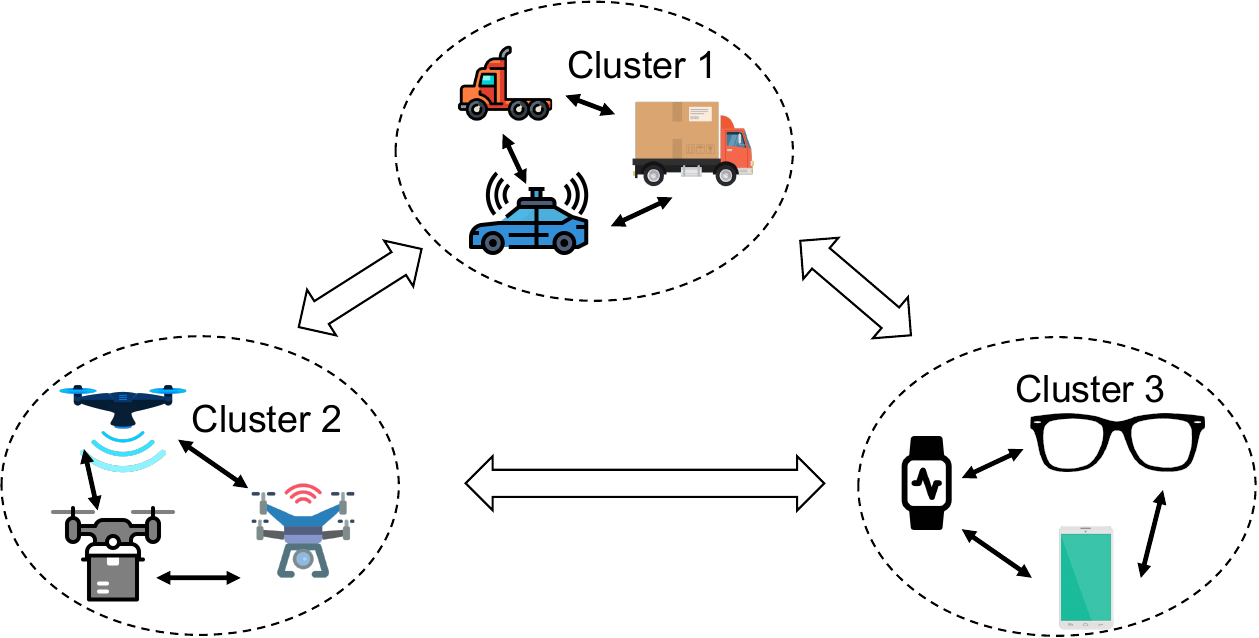}
    \caption{The clique-based topology structured in D-cliques \cite{bellet2021d} that could affect the structure of underlying topologies.}
    \Description[An example of clique-based topology]{Three clusters connected. Cluster 1 includes a forklift, a delivery truck, and a car connected. Cluster 2 contains two drones and a surveillance camera connected. Cluster 3 features a wearable smartwatch, glasses, and a smartphone connected.}
    \label{fig:d-cliques}
\end{figure}

Cliques are defined in graph theory, referring to a subset of vertices of an undirected graph such that every two distinct vertices in the clique are adjacent \cite{alba1973graph}. Cliques have been well-studied in graph theory. Cliques have also been used in FL for improving the accuracy in sparse neural networks \cite{bellet2021d}. We used the clique-based topology structure from \citet{bellet2021d} as a visualized example shown in Fig. \ref{fig:d-cliques}.

D-cliques \cite{bellet2021d} was a topology that reduced the gradient bias by grouping nodes in sparsely interconnected cliques such that the label distribution in a clique is representative of the global label distribution. This way, the impact of label distribution skew can be mitigated for heterogeneous data. Instead of providing a fully connected topology which may be unrealistic with large numbers of clients, D-Cliques instead provided locally fully connected neighborhoods. Each node belonged to a Clique, a set of fully connected nodes with data distribution as close as possible to the global distribution of the data through the network. Each Clique of the network provided a fair representation of the true data distribution, while substantially reducing the number of links. 

\subsubsection{Grid Topology}

The grid topology also enables data transmission between adjacent clients. Compared to the ring topology, where each client has two one-hope neighbors, the grid topology gives each client four neighbors, encouraging more data exchange in local networks. Shi et. al \cite{shi2022federated} discussed a scenario of distributed federated learning in a multi-hop wireless network. The experiments evaluated the performance over the line, ring, star, and grid networks, proving that more neighbors would lead to faster convergence and higher accuracy.

\subsubsection{Hybrid Topology}
In the previous section, we have introduced various types of topologies frequently seen in prior FL studies. Though these topologies cover a great portion of the use cases, each topology has its pros and cons. Researchers have explored combinations of various topologies to receive maximum benefits from network topologies. Hybrid topologies \cite{hosseinalipour2020federated, chou2021efficient, tao2021efficient, lee2020tornadoaggregate} combine the strengths of at least two traditional topologies to create a more dynamic solution.

\subsubsection{Fog Topology (Star + Mesh)}  
\begin{figure}[htb]
    \centering
    \includegraphics[width=0.65\linewidth]{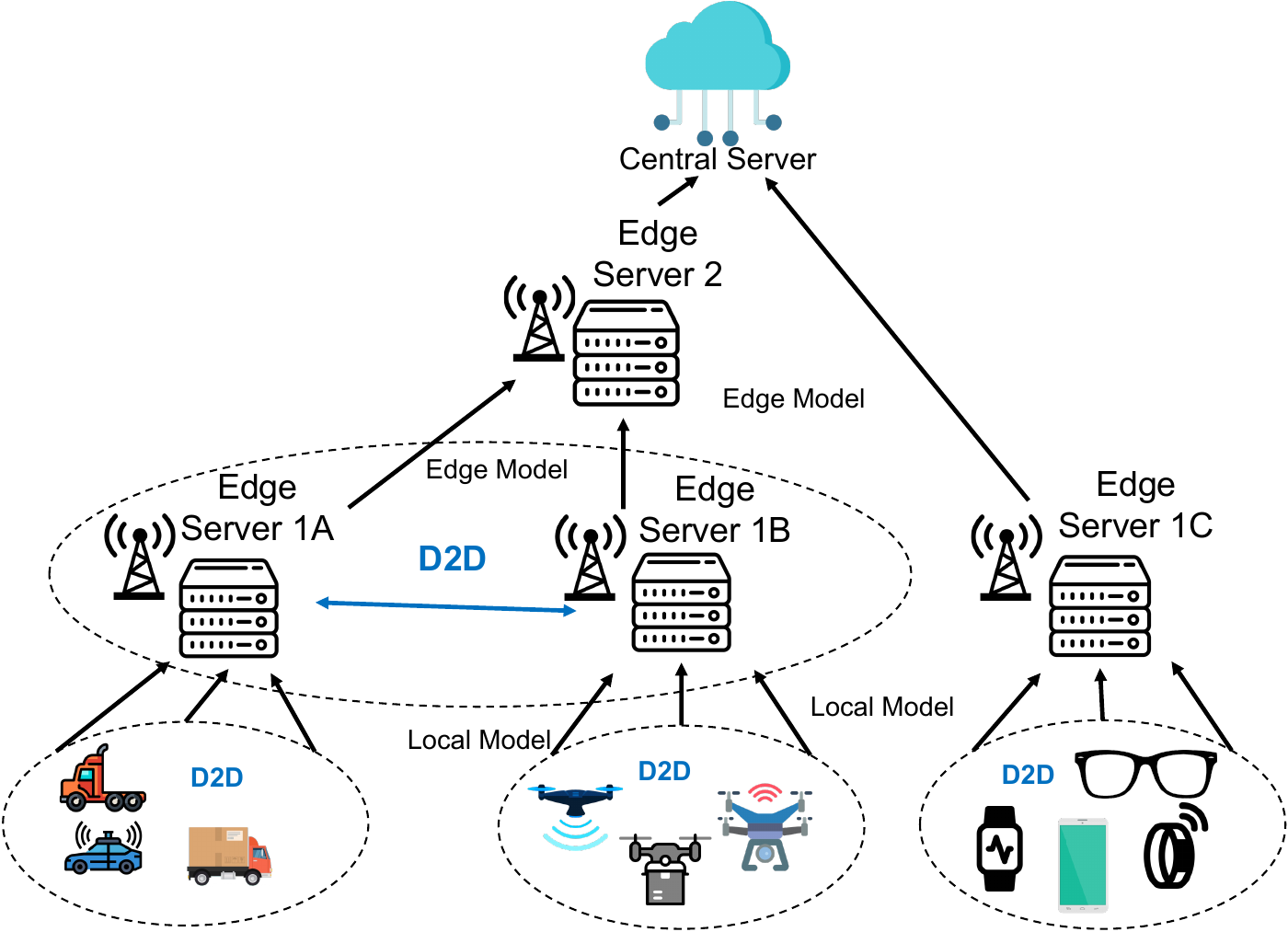}
    \caption{The fog learning topology \cite{hosseinalipour2020federated}, where D2D communications are enabled among the clients in the same cluster, as well as edge servers in the same cluster. }
    \Description[An illustration of fog learning topology]{Three lower-tier Edge Servers (1A, 1B, 1C) are connected to the central server. Each edge server has an associated edge model and is connected to local devices with its respective local models. Edge Server 1A and Edge Server 1B are connected via direct device-to-device communications, reducing latency and dependency on the central server.}
    \label{fig:fog-learning}
\end{figure}

 In this section, we examine the fog topology, which is essentially a fusion of star and mesh topologies. We show a visualization of fog topology from the works of \citet{hosseinalipour2020federated} shown in Fig. \ref{fig:fog-learning}.

The concept of fog learning was presented in \cite{hosseinalipour2020federated} compared to FL over heterogeneous wireless networks. The word ``fog'' was used to address the heterogeneity across devices. Compared to FL, fog learning considers the diversity of devices with various proximities and topology structures for scalability. The proposed fog learning boasted its multi-layer network architecture and its vertical and horizontal device communications ability. Device-to-device (D2D) communications were possible when there were fewer privacy concerns. Compared to the tree topology, additional D2D communication paths were added at the edge layer. The D2D offloading could happen among trusted devices, at the cost of privacy compromise, if such a sacrifice were acceptable. Inter-layer data offloading could also be implemented to increase the similarity of local data and reduce model bias. 

\citet{hosseinalipour2020multi} developed multi-stage hybrid federated learning (MH-FL) built on fog learning which is a hybrid of intra- and inter-layer model learning that considered the network as a multi-layer, hybrid structure with both mesh topology and tree topology. In MH-FL, each layer of which consists of multiple device clusters. MH-FL considered the topology structures among the nodes in the clusters, including local networks formed via D2D communications. It orchestrated the devices at different network layers in a collaborative/cooperative manner to form a local consensus on the model parameters and combined it with multi-stage parameters relaying between layers of the tree-shaped hierarchy. These clusters were designed in two types: limited uplink transmission (LUT) clusters with limited capability to upload data to the upper layer, and extensive uplink transmission (EUT) clusters with enough resources to perform conventional FL. 

Strictly speaking, some topologies are based on tree topologies but with additional edges \cite{zhong2021p}, making them more genetic graphs with more connectivity. To scale up FL, Parallel FL (PFL) systems were built with multiple parameter servers (PS). A parallel FL algorithm called P-FedAvg was proposed in \cite{zhong2021p}, extending FedAvg by allowing multiple parameter servers to work together. The authors identified that a single parameter server became the bottleneck due to two reasons: the difficulty in establishing a fast network that connects all devices to a single PS, as well as the limited communication capacity of only one PS. With the P-FedAvg algorithm, each client conducted several local iterations before uploading the model parameters to its PS. A PS collected model parameters from selected clients and conducted a global iteration by aggregating the model parameters uploaded from its clients and then mixing model parameters with its neighbor PS. The authors optimized the weights for PS to mix their parameters with neighbors. Essentially, this was a non-global aggregation without requiring communications with the central server. The study indicated that PFL could significantly improve the convergence rate if the network was not sparsely connected. They also compared its P-FedAvg under three different network topologies Ring, 2d-torus, and Star while 2d-torus is the most robust.

FedP2P was proposed in \cite{chou2021efficient}, aiming at reorganizing the connectivity structure to distribute both the training and communication on the edge devices by leveraging P2P communication. While edge devices performed pairwise communication in a D2D manner, a central server was still in place. However, the central server only communicated with a small number of devices. Each of these small numbers of devices represented the partition. The parameters had been aggregated before being transmitted to the central server. Compared to the tree-based topology, FedP2P was robust if one or more nodes in a P2P subnetwork were down. Compared to the original star topology, FedP2P still had better scalability with clustered P2P networks. Lin et al. proposed a semi-decentralized learning architecture called TT-HF \cite{lin2021semi}, which combined the traditional star topology of FL with decentralized D2D communications for model training, formulating a \textit{semi-decentralized} FL topology. The problem of resource-efficient federated learning across heterogeneous local datasets at the wireless edge was studied. D2D communications were enabled. A consensus mechanism to mitigate model divergence was developed to mitigate low-power communications among nearby devices. TT-HF incorporated two timescales for model training, including iterations of stochastic gradient descent at individual devices and rounds of cooperative D2D communications within clusters. 

\citet{dinh2021enabling} proposed an edge network architecture that decentralized the model aggregation process at the server and significantly reduced the aggregation latency. First, an in-network aggregation process was designed so that the majority of aggregation computations were offloaded from the cloud server to edge nodes. Then a joint routing and resource allocation optimization problem was formulated to minimize the aggregation latency for the whole system at every learning round. Numerical results showed a 4.6 times improvement in the network latency. FedCH \cite{wang2022accelerating} constructed a special cluster topology and performed hierarchical aggregation for training. FedCH arranged clients into multiple clusters based on their heterogeneous training capacities. The cluster head collected all updates from clients in that cluster for aggregation. All cluster headers took the asynchronous method for global aggregation. The authors concluded that the convergence bound was related to the number of clusters and the training epochs, and then proposed an algorithm for the optimal number of clusters with resource budgets and with the cluster topology, showing an improvement of completion time by 49.5-79.5\% and the network traffic by 57.4-80.8\%.

\begin{figure}[htb]
    \centering
    \includegraphics[width=0.65\linewidth]{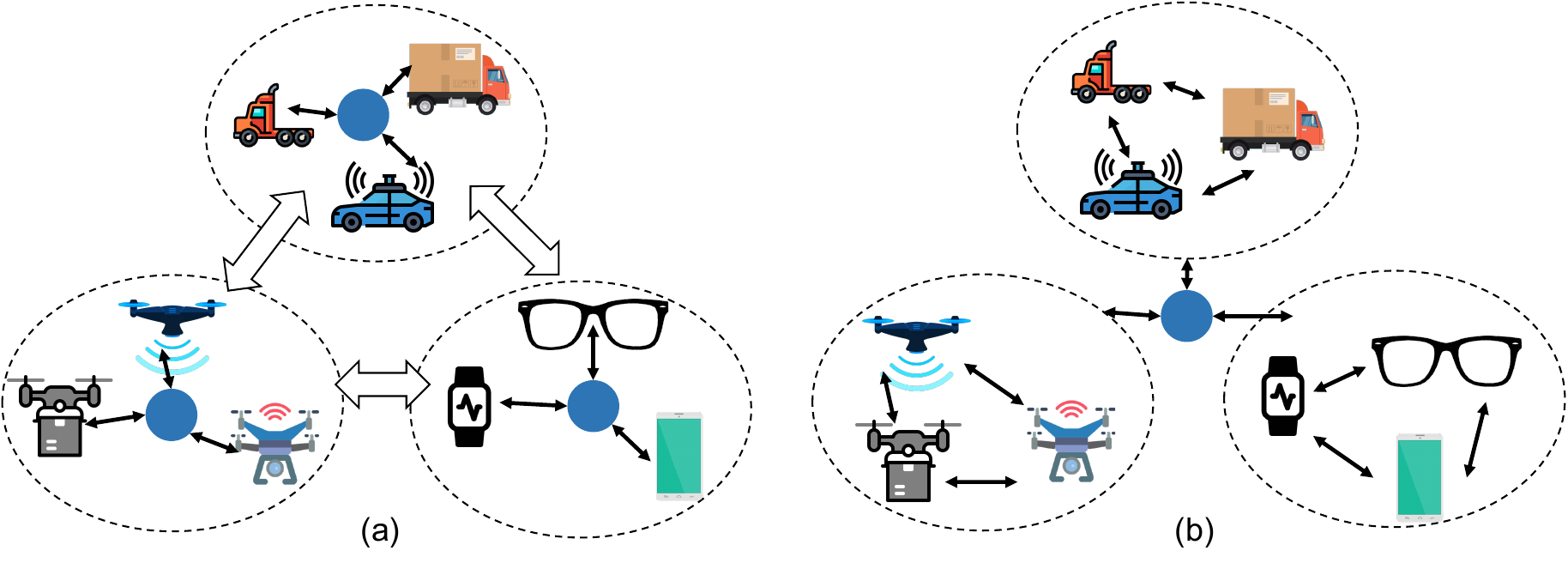}
    \caption{The semi-ring topology described in \cite{lee2020tornadoaggregate}.}
    \Description[An illustration of semi-ring topology]{There are two types of connections. The Left connects all three clusters; within each cluster, all devices are connected to an inner node.
    The right is the opposite where all devices are connected within each cluster and each cluster is connected to the same inner node. 
}
    \label{fig:semi-ring}
\end{figure}

\subsubsection{Semi-ring Topology (Ring + Star/Tree)}  
In addition to comprehensively comparing FL system structures with different topologies, the ring and tree topology were used in \cite{tao2021efficient} by Tao et al. for efficient parameter aggregation. A hybrid network topology design was proposed integrating ring (R) and n-ary tree (T) to provide flexible and adaptive convergecast in federated learning. Participating peers within one-hop were formed as a local ring to adapt to the frequent joining and leaving of devices; an n-ary convergecast tree was formed from local rings to the aggregator for communication efficiency. Theoretical Analysis found that the hybrid (R+T) convergecast design was superior for system latency. We show these hybrid topologies, termed as semi-ring Topology in Fig. \ref{fig:semi-ring}.

Lee et al. presented an algorithm called TornadoAggregate \cite{lee2020tornadoaggregate} by facilitating the ring architecture to improve the accuracy and scalability of FL. A global inter-node transfer model that is synchronized with the new model will replace traditional global aggregation in the traditional star architecture. TornadoAggregate can achieve a low convergence bond and satisfy the diurnal property condition.

\section{Challenges and Future Research Roadmaps}
\label{sec:chall}

Despite great attention addressing topology-related challenges for edge computing FL in recent years \cite{mhaisen2021optimal, wang2021efficient, bellet2021d, chen2021dacfl}, the nature of the network topologies and data distribution still introduces unique challenges. Apart from the previously mentioned topology-aware FL works, much is still to be studied about network topology in FL. In this section, we provide some open challenges and research directions for topology-aware FL.


\subsection{Topology selection}
When implementing or designing a topology-aware FL approach, there are a few things to consider. For example, in hierarchical \& heterogeneous edge networks, multiple paths exist from the edge devices to the edge servers and the central server. When selecting the topology optimal for FL, the following questions need to be asked:
\begin{enumerate}
    \item \textbf{Does a server-less architecture fit the system?}   
    \item \textbf{Is the traditional star topology no longer sufficient for the system?} 
    \item \textbf{Is there a unique topology that already exists at the hardware or structural level in the system?}
\end{enumerate}
When a system structure already exists, for example, a network of devices and subgroups of those devices controlled and managed by an intermediate server, the tree topology structure will be an obvious choice. As a result, the focus will shift from which topology to select to how to optimize the tree topology structure for a specific goal, i.e., for increased communication efficiency or to mitigate security bottlenecks. This area offers many opportunities for further research, including new topologies development or combinations of existing topologies.


\subsection{Communication cost} 
It is common for edge devices to be powered by batteries. Performing local model aggregation and wireless model transmission consume the limited power sources of edge devices. Saving communication costs and developing energy-aware federated algorithms are aligned with our primary goals. Moreover, Existing solutions save communication costs from the amount of data transmissions. Further research can be conducted on topology control algorithms that can optimize energy consumption in conjunction with network topology. 

The network heterogeneity and ever-changing nature of edge devices pose great challenges to FL.
The heterogeneity of the edge networks determines that the bandwidth resources vary for links in edge networks. Meanwhile, the mobility and density of the edge devices further reduce the actual bandwidth of those links. In the worst case, certain links in the edge networks suffer connectivity loss. The changing link conditions in edge networks demand dynamic and fault-tolerant network topologies for FL to aggregate data. Based on the amount of data for model transmission, models and algorithms are needed to find the most effective topologies to deliver the model reliably on time.

\begin{figure}[htb]
    \centering
    \includegraphics[width=0.5\linewidth]{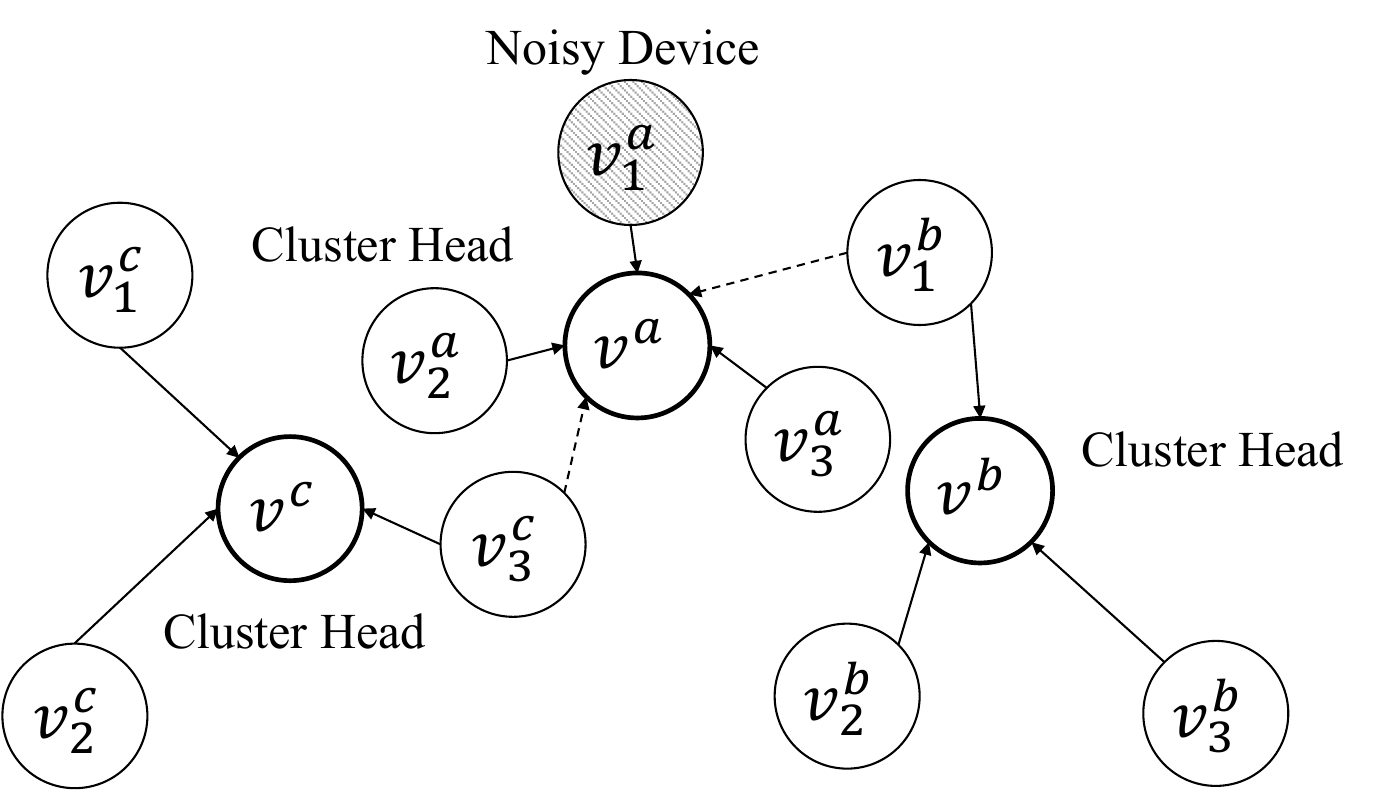}
    \caption{Topology control for cluster-based model fairness with noisy edge devices.}
    \Description[An illustration of topology control for model fairness in FL.]{The network consists of 3 Cluster Heads va, vb, and vc. A noisy device va1 exists in the network. Routing and slicing are used to reduce the exposure of the noisy device.}
    \label{fig:fl-noise}
\end{figure}


\subsection{Client-drift}
Due to the large number of edge devices and statistical heterogeneity, a phenomenon known as client drift \cite{karimireddy2020scaffold} could occur. Client drift occurs when clients with non-IID data develop extremely distinct local models away from the global optimal model. Some clients can be seen as noisy since their updates can be misleading global models. Some edge devices may become particularly ``noisy'' and their local model updates can dominate the global model weights. The problem is even more severe in the clusters with one noisy edge device. If FL leverages the cluster's local model for FL, it may be too biased towards the models by the noisy nodes. A solution to mitigate such a biased model at the cluster heads or edge servers can be an opportunistic routing that intentionally integrates models from edge devices outside a cluster. For our example in the left of Fig. \ref{fig:fl-noise}, the local models learned can be contributed to multiple clusters. The slicing of the data reduces the exposure of repeated model transmissions updated on the same dataset to the same edge servers and therefore enhances privacy.


\subsection{Ethical/Privacy concerns}
In this section, we discuss new ethical and privacy challenges possessed by different network topology structures. The primary concern with the standard star topology is the communication bottleneck and excessive reliance on a central server. The central server is heavily tasked with safeguarding all client information. The default star topology in FL can be represented as a single point of failure potentially compromising the privacy of the entire network and raising potential ethical concerns. Other network structures can address some of the privacy concerns of the traditional topology structure. 
For tree topology, the additional communication layers and intermediate servers present both advantages and disadvantages. The tree topology allows the central server to offload some computational tasks and client information to the intermediate server. However, the presence of the intermediate servers requires greater efforts towards privacy on the edge. Other fully decentralized topologies do not require a central server, such as mesh, gossip, clique, and grid topology. Without an overarching central server, there is no need for direct communication with the server, which would normally pose a significant privacy threat and ethical violations. What comes with this is the increased amount of peer-to-peer (P2P) communication which could introduce new privacy challenges.

The extended period of model aggregations, ``devices -> edge servers -> central server'', could lead to privacy concerns regarding the training data of the edge devices. 
When an edge device sends its aggregated model upon a global model update request, the model will be broadcast to its neighbors. Repeated rounds of sharing models among the nearby neighbors will mix the device's local model with the sub-local models. The changes in the topology select changing sets of neighbors, further increasing the diversity of local models. The model mingling activities will reduce the vulnerability of devices suffering differential privacy attacks.

In conclusion, privacy is always a trade-off. No single topology can meet all the needs of all users. As technology advances, the existing topology will face new challenges. New and niche topologies will present new challenges and opportunities. Various aspects of network topology and their impact on ethical and privacy issues require further research.

\subsection{Availability-aware FL assisted by topology overlay}
\begin{figure}[htb]
    \centering
    \includegraphics[width=0.36\linewidth]{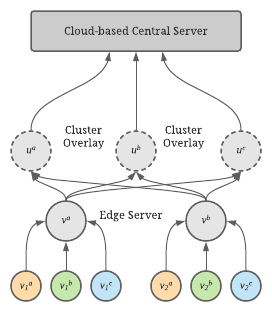}
    \caption{An overlay network on top of edge computing clusters.}
    \Description[An illustration of overlay network]{Cloud-based Central Server is on the highest level. There are three cluster overlays below and build on top of the physical edge server to optimize the management and efficiency of tasks across the edge network.}
    \label{fig:fl-overlay}
\end{figure}

On top of that, the model aggregation tasks at the edge servers, also known as cluster heads in a hierarchical FL architecture, face availability challenges when the edge servers are down. While with the meshed links among the edge servers, learning task replication techniques can be used for maintaining the availability levels of the model aggregation tasks. In other words, we must make a trade-off between robustness and redundancy. The problem can be further investigated from a resource allocation perspective, scheduling, and clustering. In the right of Fig. \ref{fig:fl-overlay}, the clusters can be built logically instead of following the physical topology of the edge networks based on the intensity and the distribution of data generation. In the example, there are two physical clusters $v^a$ and $v^b$. The three edge devices in a cluster belong to one of the three separate overlay clusters $u^a$, $u^b$, and $u^c$.


\subsection{Conduct real-world deployment}
In most topology-related FL works, their experiments are conducted in simulated environments, with the exceptions of \cite{hard2018federated, zhou2020privacy, pokhrel2020federated, tran2018joint, dinh2021enabling, wang2019edge, khan2023outdoor, topham2022gait}. Within these works, \citet{zhou2020privacy} only used the Alibaba Cloud as its parameter server and \citet{tran2018joint} simply designed their experiment using realistic model settings from \cite{chu2013heterogeneous}. The experiment setting for \cite{dinh2021enabling} although still simulations, does involve real deployment on a grid network inside a 500m×500m area. \citet{wang2019edge} use a unique approach that captures the \textit{Xender’s} trace content and requests files from active mobile users. Many FL works to date do not include or consider real-world model deployments in their experiments. However, this type of work can demonstrate real challenges faced when deploying unique edge topologies in a realistic setting while tackling specific issues such as model deployment time, inference time, communication costs, etc. For further research, it would be advantageous to conduct experiments with real-world deployment and evaluate topology-aware FL studies in real edge environments. This approach ensures the proposed FL techniques and algorithms can be properly validated, moving beyond just proofs-of-concept in simulated environments. For topology-aware FL, much work can be done to develop a realistic test bed and to perform real-world deployments.

\section{Conclusion}
\label{sec:conclu}
In this survey, the role of topology-aware federated learning in edge computing is discussed in detail. Various network topologies, including star, tree, decentralized, and hybrid topologies, are summarized and compared to illustrate the substantial impact of topology on the efficiency and effectiveness of federated learning. Different topologies can bring many benefits to the network. It is important to note that various topology structures will undoubtedly bring extra complexity. There is a choice to be made if the simple star topology cannot meet the needs of a growing system and infrastructure and whether to opt for another topology for increased communication and complexity. FL architectures must also account for factors such as the central server's necessity or absence, clients' diurnal activity patterns, and options for implementing intermediate servers.



\bibliography{main}

\end{document}